\documentclass[10pt,twocolumn,letterpaper]{article}

\usepackage[pagenumbers]{iccv} 
\usepackage{pifont}
\definecolor{iccvblue}{rgb}{0.21,0.49,0.74}
\usepackage[pagebackref,breaklinks,colorlinks,allcolors=iccvblue]{hyperref}

\def\model{SMQ} 

\def\paperID{8621} 
\def\confName{ICCV}
\def\confYear{2025}

\title{Skeleton Motion Words for Unsupervised Skeleton-Based Temporal Action Segmentation}

\author{
Uzay G\"okay$^{1}$ \quad Federico Spurio$^{1,3}$ \quad Dominik R. Bach$^{1,2}$ \quad Juergen Gall$^{1,3}$ \\
$^{1}$University of Bonn, Germany \quad
$^{2}$University College London, UK\\
$^{3}$Lamarr Institute for Machine Learning and Artificial Intelligence, Germany 
\\{\tt\small \{u.goekay, d.bach\}@uni-bonn.de \quad \{fspurio, gall\}@iai.uni-bonn.de}
}

\begin{document}
\maketitle
\begin{abstract}
Current state-of-the-art methods for skeleton-based temporal action segmentation are predominantly supervised and require annotated data, which is expensive to collect. In contrast, existing unsupervised temporal action segmentation methods have focused primarily on video data, while skeleton sequences remain underexplored, despite their relevance to real-world applications, robustness, and privacy-preserving nature. 
In this paper, we propose a novel approach for unsupervised skeleton-based temporal action segmentation. Our method utilizes a sequence-to-sequence temporal autoencoder that keeps the information of the different joints disentangled in the embedding space. Latent skeleton sequences are then divided into non-overlapping patches and quantized to obtain distinctive skeleton motion words, driving the discovery of semantically meaningful action clusters. We thoroughly evaluate the proposed approach on three widely used skeleton-based datasets, namely HuGaDB, LARa, and BABEL. The results demonstrate that our model outperforms the current state-of-the-art unsupervised temporal action segmentation methods. Code is available at \href{https://github.com/bachlab/SMQ}{\texttt{github.com/bachlab/SMQ}}.
\end{abstract}
    
\section{Introduction}
\label{sec:intro}

Advancements in motion capture technology and pose estimation have enabled the precise recording of human motion, represented as time series of joint positions, orientations, or other representations of skeleton sequences. Recently, there has been an increased interest in temporally segmenting long skeleton sequences into action segments~\cite{filtjens2022skeleton, LaSA_2024ECCV, xu2023efficient, hosseini2020deep, tan2023hierarchical, li2023decoupled, tian2024spatial, tian2023stga}. These approaches, however, are predominantly supervised, requiring extensive labeled data for training. In contrast, prior unsupervised skeleton-based approaches~\cite{su2020predict, lin2020ms2l, zhang2022contrastive, guo2022contrastive, lin2023actionlet, li20213d, paoletti2022unsupervised} focus mainly on action recognition, assuming that the sequences are very short and contain only a single action per sequence. This highly constrained setting limits their applicability to real-world scenarios, where multiple actions are performed within a single sequence.

In the context of RGB videos, unsupervised approaches for temporal action segmentation have been proposed~\cite{kukleva2019unsupervised, li2021action, kumar2022unsupervised, lin2023taec, xu2024temporally, tran2024permutation, spurio2025hvq}. These methods discover actions in a set of long untrimmed video sequences by identifying actions across sequences and multiple actions within each sequence. While these approaches achieved promising results on video data, they have not been evaluated on skeleton sequences. Our investigations reveal that such methods either struggle to recognize re-occurring actions due to post-processing with Viterbi decoding or produce overly small segments, as visualized in qualitative results.

\begin{figure}[t]
    \centering
    \includegraphics[width=\columnwidth]{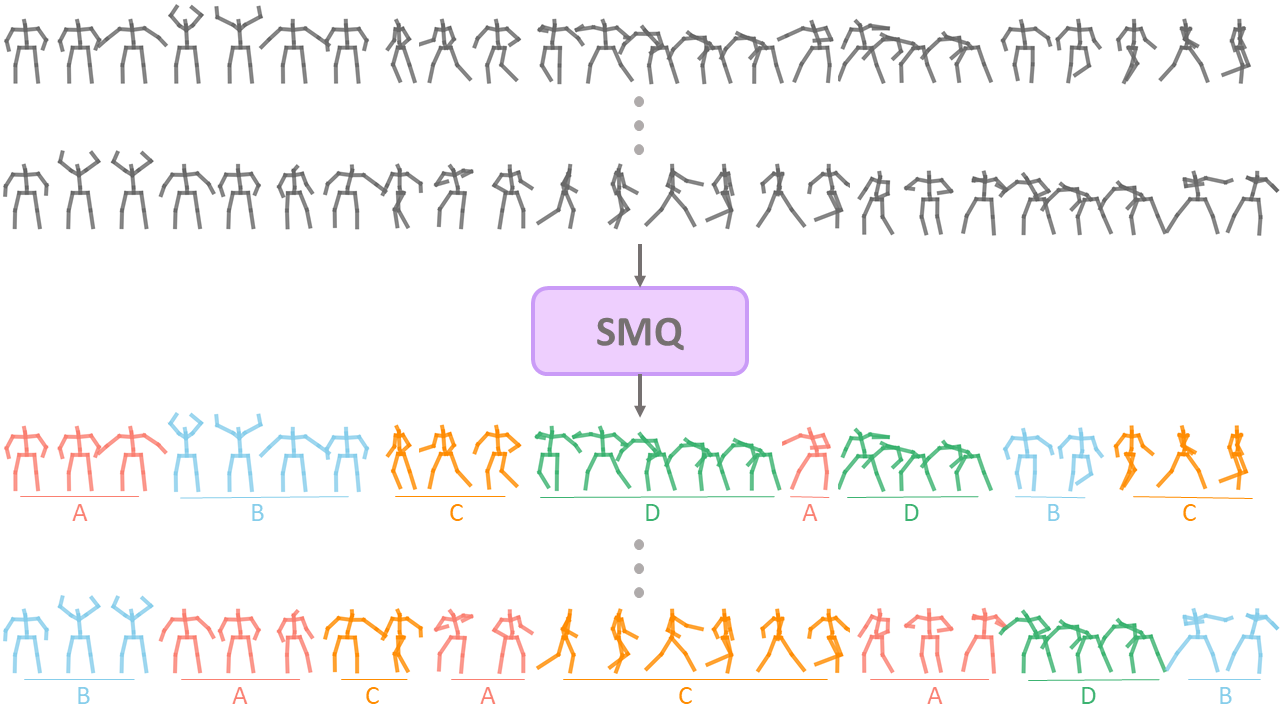}
    \caption{Our unsupervised skeleton-based temporal action segmentation framework, Skeleton Motion Quantization (\textit{\model}), takes a set of untrimmed skeleton sequences as input (top) and discovers actions consistent across all sequences (bottom). This is achieved by jointly segmenting and clustering all sequences. The letters A, B, C, and D correspond to the identified actions.
    }\vspace{-4mm}
    \label{fig:skeleton-tas}
\end{figure}

In this paper, we thus propose a novel approach for unsupervised skeleton-based temporal action segmentation, which is illustrated in Fig.~\ref{fig:skeleton-tas}. Our model uses a dilated temporal convolution based sequence-to-sequence autoencoder, which preserves the skeleton structure by ensuring that the information of the different joints remains disentangled in the embedding space. We will show that this joint-based disentanglement is essential for improving performance. Subsequently, the latent representations generated by the temporal encoder are then patchified into non-overlapping sub-sequences of motion, which are quantized by a skeleton motion quantization module. Specifically, each latent motion patch is assigned to the closest cluster prototype, and these prototypes are learned during training with the aim of capturing the most representative skeleton motion words of each action in the dataset. We term the approach thus \textbf{S}keleton \textbf{M}otion \textbf{Q}uantization \textbf{(SMQ)}. 

To assess the effectiveness of \textit{\model}, we evaluated it on three skeleton-based datasets, namely HuGaDB~\cite{chereshnev2018hugadb}, LARa~\cite{niemann2020lara}, and BABEL~\cite{BABEL:CVPR:2021}. We demonstrate that it outperforms state-of-the-art approaches for unsupervised temporal action segmentation by a large margin, which highlights the importance of developing methods that exploit the inherent structure of skeleton sequences. We also show that it outperforms existing skeleton-based representation learning approaches on the task of unsupervised skeleton-based temporal action segmentation.   
In summary, our contributions are the following: 
\begin{enumerate}
    \item We thoroughly evaluate existing unsupervised temporal action segmentation approaches for skeleton data and provide protocols for three datasets.
    \item We introduce \textit{\model}, a novel unsupervised framework for skeleton-based temporal action segmentation. It learns representative skeleton motion words in a joint-disentangled embedding space.
    \item We demonstrate that \textit{\model} outperforms existing unsupervised action segmentation methods, underscoring the importance of utilizing the structure of skeleton data.
\end{enumerate}
\section{Related Work}
\label{sec:related_work}

\textbf{Human Motion Segmentation.}
Human motion segmentation aims to break down continuous motion into distinct, meaningful segments, often at a very fine granularity \cite{barbivc2004segmenting, zhouaca, kruger2016efficient, sedmidubsky2020motion}, such as phases of a gait cycle. In contrast, temporal action segmentation targets broader, higher-level actions over long, untrimmed sequences~\cite{ding2023temporal}. One of the earliest approaches for human motion segmentation was introduced by Barbi{\v{c}} \etal~\cite{barbivc2004segmenting}, using a probabilistic PCA-based method to detect significant distribution changes in human poses. Other early methods include hierarchical clustering and graph-based techniques to align sequences and segment motion based on structural similarity \cite{zhou2012hierarchical, zhouaca, kruger2016efficient, vogele2014efficient}. Recent methods for human motion segmentation use heuristics such as extended Douglas-Peucker keypoint detection~\cite{ma2021fine}, zero-velocity crossing with adaptive clustering~\cite{wang2015unsupervised}, and curvature analysis on kinematic data~\cite{arn2018motion} to isolate motion primitives within individual sequences. While these approaches focus on fine-grained motion patterns, there are recent methods \cite{sarfraz2021tw-finch, du2022abd, bueno2023tsa} that aim to identify complex, high-level actions within individual video sequences. However, they are still limited to segmenting a single sequence at a time. Our approach, in contrast, identifies globally learned actions across multiple sequences and subjects.

\textbf{Temporal Action Segmentation.}
Temporal action segmentation involves segmenting long untrimmed videos, often containing multiple actions per video, into contiguous segments, each labeled with a specific action~\cite{kuehne2016end, lea2017temporal, farha2019ms, huang2020improving, behrmann2022unified, liu2023diffusion, lu2024fact}. Unlike traditional action recognition, which aims to classify short videos containing a single action, temporal action segmentation addresses the sequential nature of actions. 
To avoid an expensive annotation process, unsupervised methods \cite{sener2018unsupervised, kukleva2019unsupervised, li2021action, kumar2022unsupervised, lin2023taec, xu2024temporally, tran2024permutation, spurio2025hvq} have been proposed for this task.
Specifically, Mallow~\cite{sener2018unsupervised} is a two-stage iterative approach, combining discriminative learning from visual features with generative modeling of temporal structure using a Generalized Mallows Model. CTE~\cite{kukleva2019unsupervised} aims to learn continuous temporal embeddings with the supervision of the relative time-stamps and then clusters the learned features in a two step fashion. TOT~\cite{kumar2022unsupervised} proposed a joint representation learning and online clustering framework which applies temporal optimal transport on the features from the encoder and cluster prototypes to obtain pseudo action labels for training. UFSA~\cite{tran2024permutation} combines the encoder from ASFormer~\cite{yi2021asformertransformeractionsegmentation} and the decoder from UVAST~\cite{behrmann2022unified}, allowing the use of segment-level cues. Lastly, ASOT~\cite{xu2024temporally} utilizes temporally consistent unbalanced optimal transport, relaxing the balanced assignment assumption in TOT. Our results demonstrate that despite progress in unsupervised temporal action segmentation, existing methods do not perform well on skeleton data, as they do not exploit the inherent structure of skeleton data. 

\textbf{Skeleton-based Temporal Action Segmentation.}
Specifically for skeleton data, a variety of fully supervised temporal action segmentation methods have been proposed. \cite{filtjens2022skeleton, xu2023efficient, tian2023stga, yan2018spatial} combine GCNs \cite{kipf2017semisupervised} and TCNs \cite{lea2017temporal,farha2019ms,9186840} to capture both the spatial and temporal aspects of these sequences. Additionally, some skeleton-based temporal action segmentation approaches further investigate the separation \cite{li2023involving} and decoupling \cite{li2023decoupled} of spatial and temporal information. Various attention mechanisms \cite{liu2022spatial, tian2023stga} have been proposed to capture spatio-temporal dependencies while other approaches emphasize distinct frameworks, including action synthesis \cite{xu2023efficient} and language priors \cite{LaSA_2024ECCV}. However, all of these methods rely on full supervision, requiring extensive labeling. 
In this work, we propose a skeleton-based temporal action segmentation approach that automatically learns holistic, globally consistent actions across the entire dataset in an unsupervised manner.

\textbf{Self-Supervised Representation Learning for Skeleton Data.}
Self-supervised representation learning \cite{chen2020simple, he2020momentum, grill2020bootstrap, caron2018deep, caron2020unsupervised, caron2021emerging, he2022masked, feichtenhofer2022masked} aims to learn feature representations without human annotations by solving pretext tasks that exploit the structure of unlabeled data. Recent methods have explored this paradigm for skeleton-based action recognition ~\cite{su2020predict, xu2021unsupervised, zhang2022contrastive, guo2022contrastive, lin2023actionlet, li20213d, zheng2018unsupervised, lin2020ms2l}, focusing on short, trimmed sequences containing a single action. In particular, CrosSCLR~\cite{li20213d}, AimCLR~\cite{guo2022contrastive}, and ActCLR~\cite{lin2023actionlet} adopt contrastive learning to extract discriminative representations. However, these methods solely learn representations without predicting action labels during training, operating on short, single-action sequences. They thus require annotations for downstream evaluation and are not applicable to long, untrimmed sequences.

hBehaveMAE~\cite{Stoffl2024.08.06.606796} uses a hierarchical framework with a masked autoencoder to learn interpretable latent representations of actions at varying levels of granularity. LAC~\cite{yang2023lac} utilizes pre-trained visual encoders with the task of learning composable actions from synthesized data. The model is then fine-tuned for different tasks including action segmentation to evaluate the learned representation. Unlike self-supervised approaches that learn action-related representations without assigning actions during training, and therefore require annotated data for downstream fine-tuning, our method discovers and segments actions directly during training in a fully unsupervised and label-free manner.
\begin{figure*}[htbp]
    \centering
    \includegraphics[trim=0pt 140pt 155pt 0pt, clip,width=0.9\linewidth]{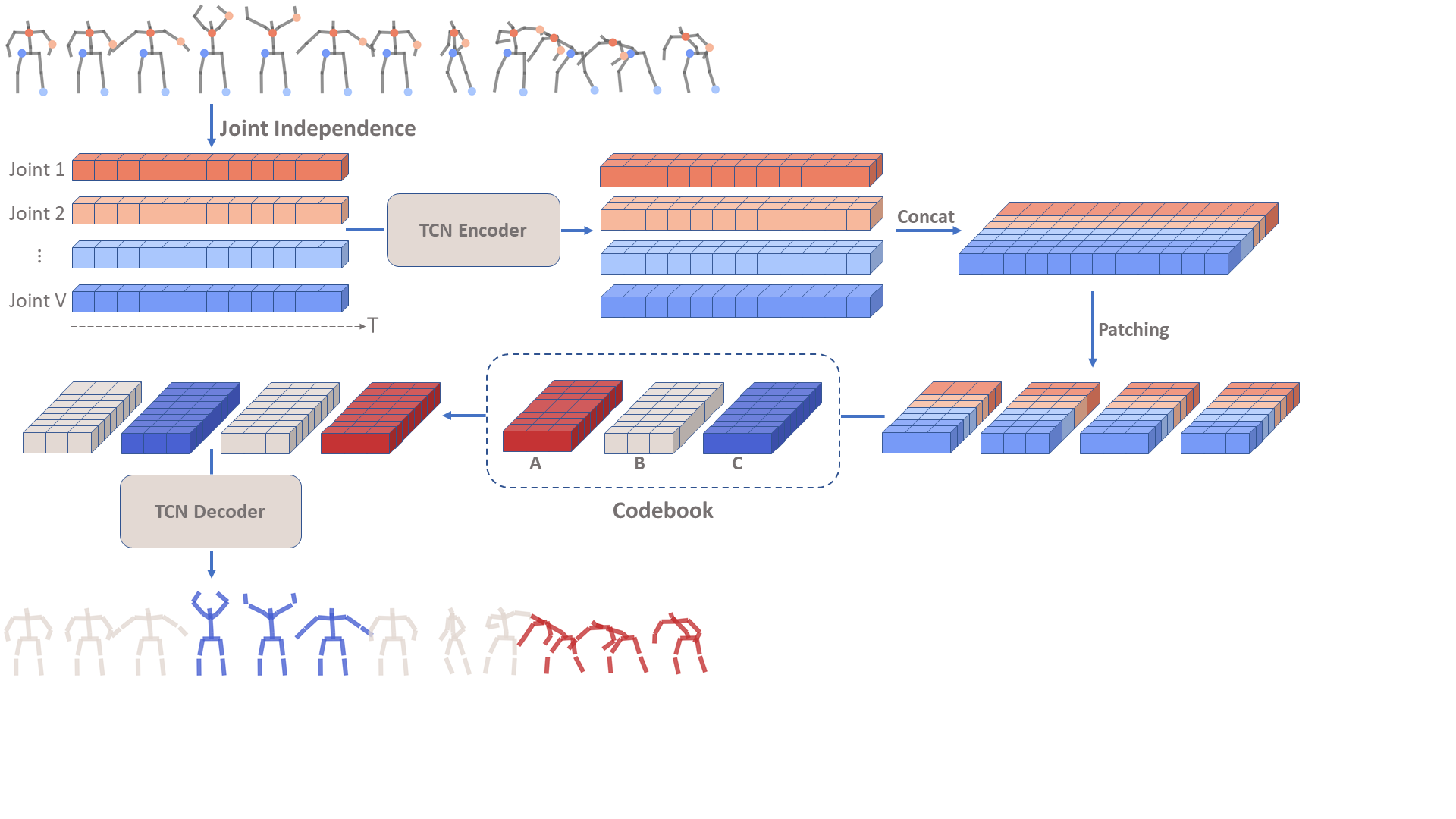}
    \caption{Overview of \textit{\model}. Given a set of skeleton sequences, our approach discovers actions that are semantically consistent across all sequences. The discovered actions correspond to clusters of learned motion words in the codebook. The model first encodes the skeleton sequences into a joint-based disentangled embedding space. Each embedded sequence is then divided into short non-overlapping temporal patches and a codebook with $K$ motion words is learned for the entire dataset via a patch-based quantization process, which assigns each patch to its nearest motion word using Euclidean distance. This assignment directly provides the segmentation of each sequence based on the motion word indices. In order to learn meaningful representations, the decoder reconstructs the input sequences from the quantized patches, where each patch is replaced by its corresponding motion word.}
    \vspace{-4mm}
    \label{fig:smq-teaser}
\end{figure*}

\section{Unsupervised Skeleton-based Temporal Action Segmentation}\label{sec:method}
In this work, we address the task of unsupervised skeleton-based temporal action segmentation, which is illustrated in Fig.~\ref{fig:skeleton-tas}. Given input skeleton sequences encoding joint positions, orientations, or related features, the objective of skeleton-based action segmentation is to map each frame to one of the predefined action categories. In the unsupervised case, the actions are unknown and need to be discovered from the data. Formally, the input skeleton sequences are defined as $\mathbf{X} \in \mathbb{R}^{N \times C \times T \times V}$, where $N$ is the number of sequences or batch size, $C$ is the number of feature channels representing a single joint, $T$ is the sequence length, and $V$ is the number of joints. The goal is to discover \( K \)  semantically meaningful clusters and to generate an output label sequence \( \mathbf{Y} \in \mathbb{N}^{N \times T} \), where each element \( \mathbf{Y}_{nt} \in \{1, 2, \dots, K\} \) corresponds to the cluster assigned to the \( t \)-th frame of the \( n \)-th sequence. The main challenge is to discover actions across sequences and subjects, since it requires dealing with a large variability in how actions can be executed by different individuals.    

While the task has been addressed for video data, it has not been studied for skeleton data. In our experiments, we show that state-of-the-art unsupervised approaches for video-based temporal action segmentation do not perform well for skeleton data since they do not exploit the nature of skeleton data. We therefore propose an approach for unsupervised skeleton-based temporal action segmentation. 

The approach is illustrated in Fig.~\ref{fig:smq-teaser}. Each joint of a sequence will be mapped to an embedding space by an encoder,  which will be described in Section \ref{sec:encoder}. The encoder learns an embedding for each joint separately, which ensures that the information of the different joints remains disentangled in the embedding space and that the representation is not dominated by a subset of joints. Another important aspect of the approach is the conversion of a frame-wise representation into a patch-based representation, which is then quantized. We term this approach skeleton motion quantization (\textit{\model}) and describe it in Section~\ref{sec:smq}. 
Finally, the decoder reconstructs the original skeleton sequence from the quantized patches as described in Section \ref{sec:decoder}. While the end-to-end training is described in Section~\ref{sec:loss}, the following sections provide a detailed description of each component.

\subsection{Encoder}\label{sec:encoder}
In our approach, we process input skeleton sequences to capture joint-specific temporal dynamics effectively. 
To learn latent subspaces for each joint individually, we reshape and permute $\mathbf{X}$ into $\mathbf{X}' \in \mathbb{R}^{(N \cdot V) \times C \times T}$.
This transformation treats each joint's temporal sequence $\mathbf{X}_{nv}' \in \mathbb{R}^{C \times T}$ as an independent sample, allowing the encoder to capture fine-grained motion patterns unique to each joint. Specifically, the encoder is a variant of a Temporal Convolutional Network (TCN) \cite{farha2019ms,9186840}, which consists of a series of dilated residual layers that progressively refine the temporal features of $\mathbf{X}'$. Each stage begins with a \(1 \times 1\) convolution to project the input features into a higher-dimensional space, followed by several residual blocks with dilated convolutions to capture long-range temporal dependencies. The dilation factor in these residual layers increases exponentially, allowing the network to model temporal patterns at multiple scales. Finally, a \(1 \times 1\) convolutional layer maps the features to the latent dimension  ${D}$, yielding latent embeddings $\mathbf{Z}_{nv} \in \mathbb{R}^{D \times T}$.

\subsection{Skeleton Motion Words}\label{sec:smq} 
Since skeleton sequences represent joint positions, orientations, or higher-order representations as time series, we reformulate the unsupervised skeleton-based temporal action segmentation task as an online time series clustering problem \cite{van2017vqvae}. In this formulation, a skeleton sequence is mapped to a sequence of \textbf{skeleton motion words} which are learnable discrete units representing prototypical short-term temporal skeleton motion patterns. This allows us to capture the dynamic and temporal dependencies inherent in human motion data, aligning well with the nature of sequential joint movements across time.

As shown in Fig.~\ref{fig:smq-teaser}, the skeletons are first mapped into an embedding space using a temporal encoder, while keeping the information of the different joints disentangled. To model skeleton motion, the embeddings are divided into non-overlapping patches to capture local temporal structure over fixed intervals. During training, skeleton motion words are learned, and each patch is assigned to its closest motion word based on the learned representation. Therefore, it effectively captures both spatial and temporal patterns in skeleton sequences, allowing for more robust and coherent action segmentation.

\subsubsection{Temporal Patching}\label{sec:patching}
In order to achieve skeleton motion quantization, we start by patching the latent representations from the encoder into fixed-length short time series. Specifically, after reorganizing the per-joint embeddings \( \mathbf{Z}_{nv} \) and concatenating joint features of the embedded skeleton to form \( \mathbf{Z}_{concat} \in \mathbb{R}^{N \times T \times (V \cdot D)} \), we ensure that the sequence length \( T \) is divisible by the patch size \( P \) by applying zero-padding as necessary. We then partition the sequence into \( M = \frac{T}{P} \) non-overlapping temporal patches, yielding a patched tensor \( \mathbf{Z}_p \in \mathbb{R}^{N \times M \times P \times (V \cdot D)} \).

This structured patching allows us to measure similarity based on sequences of embedding vectors rather than individual vectors, accommodating greater temporal variability. This enables the learning of representative skeleton motion words that encapsulate temporal dynamics, rather than relying on single-frame representations.

\subsubsection{Quantization}\label{sec:quantization}
Based on the temporal patching \( \mathbf{Z}_p \), the network learns a patch-based codebook \( \mathcal{C} = \{\mathbf{c}_k\}_{k=1}^{K} \), where each codebook element \( \mathbf{c}_k \in \mathbb{R}^{P \times (V \cdot D)} \).
As before, \( P \) denotes the patch size, \( V \) the number of joints, and \( D \) the latent dimension.
Each element \( \mathbf{c}_k \) in the codebook defines what we refer to as a skeleton motion word, which serves as a discrete representation of skeleton motion over a temporal patch. The codebook \( \mathcal{C} \) is initialized randomly using Kaiming uniform initialization.

During quantization, each patch in the batch \( \mathbf{p}_i \in \mathbf{Z}_p \) is assigned to its nearest motion word in the codebook based on the Euclidean distance. Each patch is then replaced with its closest motion word \( \mathbf{c}_{k_i} \), where \( \mathbf{c}_{k_i} \) denotes the codebook element assigned to \( \mathbf{p}_i \), yielding the quantized patch \( \mathbf{q}_i \in \mathbb{R}^{P \times (V \cdot D)} \), \ie,
\begin{align}
\mathbf{q}_{i}=\mathbf{c}_{k_i} \quad \text{with} \quad k_i = \operatorname*{arg\,min}_{k} \, \|\mathbf{p}_i - \mathbf{c}_k\|_2\,.
\end{align}
This produces the quantized tensor \( \mathbf{Q}_p \in \mathbb{R}^{N \times M \times P \times (V \cdot D)} \). We then depatchify the quantized tensor \( \mathbf{Q}_p \) to obtain \( \mathbf{Q} \in \mathbb{R}^{N \times T \times (V \cdot  D)} \). From the assignment of a patch \( \mathbf{p}_i \) to the motion word \( \mathbf{c}_{k_i} \), we directly derive the segmentation \( \mathbf{Y} \in  \mathbb{N}^{N \times T} \) of all sequences in the batch, where each element \( \mathbf{Y}_{nt} \in \{1, 2, \dots, K\} \) denotes the index of the assigned motion word at the \( t \)-th frame of the \( n \)-th sequence.

\subsubsection{Codebook Update}\label{sec:codebook_update}
Following the quantization process, where patches are assigned to their nearest motion words, we update the motion words by averaging the assigned patches:
\begin{equation}
\mathbf{c}_k = \frac{1}{|\mathcal{P}_k|} \sum_{\mathbf{p}_i \in \mathcal{P}_k} \mathbf{p}_i, \quad \forall k \in \{1, \dots, K\},
\label{eq:codeword_update}
\end{equation}
where \( \mathcal{P}_k = \{\mathbf{p}_i \mid k_i = k\} \) denotes the set of patches assigned to the \( k \)-th motion word. To enhance the stability of codebook updates within the \textit{\model} framework, we use Exponential Moving Average (EMA) updates~\cite{van2017vqvae}:
\begin{align}
\mathbf{c}_k^{(t)} &\leftarrow \alpha \mathbf{c}_k^{(t-1)} + (1 - \alpha) \mathbf{\tilde{c}}_k^{(t)}, \label{eq:ema_update} \\
\mathbf{\tilde{c}}_k^{(t)} &= \frac{1}{|\mathcal{P}_k^{(t)}|} \sum_{\mathbf{p}_i \in \mathcal{P}_k^{(t)}} \mathbf{p}_i, \quad \forall k \in \{1, \dots, K\},
\label{eq:ema_compute}
\end{align}
where \( \mathcal{P}_k^{(t)} = \{\mathbf{p}_i \mid k_i = k\} \) is the set of patches assigned to the \( k \)-th motion word at iteration \( t \), and $\alpha=0.5$ is the EMA decay factor that controls the smoothness of the updates.

\subsection{Decoder}\label{sec:decoder}
To reconstruct the input skeleton sequences from the quantized latent representations, we design a decoder that mirrors the encoder's architecture. After quantization, the depatchified tensor \(\mathbf{Q} \) is reshaped into \(\mathbf{Q}' \in \mathbb{R}^{(N \cdot V) \times D \times T}\) to obtain the joint independence. 
This reshaping allows the decoder to process each joint's temporal sequence individually, similar to the encoder's approach. The reconstructed skeleton, denoted as \(\hat{\mathbf{X}} \in \mathbb{R}^{N \times C \times T \times V}\), is then obtained after reshaping.

\subsection{Training Loss}\label{sec:loss}
The network is trained end-to-end using two loss functions. The first term is the reconstruction loss that compares the reconstructed sequences \(\hat{\mathbf{X}} \in \mathbb{R}^{N \times C \times T \times V}\) with the original sequences $\mathbf{X} \in \mathbb{R}^{N \times C \times T \times V}$. The loss ensures that the encoder and codebook preserve as much information from the input sequences as possible. While the mean squared error (MSE) between $\mathbf{X}$ and $\hat{\mathbf{X}}$ is an obvious choice, we propose to compute inter-joint distance MSE loss, defined over the distance from each joint to all other joints within the same frame:

\begin{align}
\nonumber L_{\text{rec}} =& \frac{1}{N \cdot T \cdot V^2} \sum_{n=1}^N \sum_{t=1}^T \sum_{v=1}^V \sum_{w=1}^V \left( dX_{ntvw} - d\hat{X}_{ntvw} \right)^2\\
&\text{with}\quad
\begin{array}{ll}
dX_{ntvw} &= \left\| \mathbf{X}_{ntv} - \mathbf{X}_{ntw} \right\|_2,\\
d\hat{X}_{ntvw} &= \left\| \hat{\mathbf{X}}_{ntv} - \hat{\mathbf{X}}_{ntw} \right\|_2.
\end{array}
\end{align}

The loss has the advantage that it is inherently translation and rotation-invariant and only measures differences in poses. In our experiments, we show that the proposed loss performs better than a standard MSE loss.

To ensure that the learned embeddings are separated into distinct skeleton motion words, 
we employ the commitment loss~\cite{van2017vqvae}. This loss encourages the encoder to produce latent embeddings that are close to their assigned motion words. The commitment loss is defined as
\begin{equation}
L_{\text{commit}} = \sum_{\mathbf{p}_i \in \mathbf{Z}_p} \left\| \text{sg}\left[ \mathbf{c}_{k_i} \right] - \mathbf{p}_i \right\|_2^2 ,
\end{equation}
where \( \mathbf{c}_{k_i} \in \{ \mathbf{c}_1, \ldots, \mathbf{c}_K \} \) is the motion word that the latent patch embedding $\mathbf{p}_i$ has been assigned to, and \( \text{sg}[\cdot] \) denotes the stop-gradient operator, which stops the backpropagation of the gradient. 

The total loss function combines both the reconstruction loss and the commitment loss
\begin{equation}\label{eq:total-loss}
L_{\text{total}} = \lambda L_{\text{rec}} + L_{\text{commit}} ,
\end{equation}
where \( \lambda=0.001 \) is the hyperparameter that balances the loss terms.
\section{Experiments}

\captionsetup[table]{skip=5pt}
\begin{table*}[htb]
    \centering
    \begin{tabular}{@{}l|l|lllll|lllll@{}}
        \hline
        &  & \multicolumn{5}{|c|}{\textbf{HuGaDB}} & \multicolumn{5}{c}{\textbf{LARa}} \\
        \cline{3-12}
        & \textbf{Method} & \textbf{MoF} & \textbf{Edit} & \multicolumn{3}{c|}{\textbf{F1@\{10, 25, 50\}}} & \textbf{MoF} & \textbf{Edit} & \multicolumn{3}{c}{\textbf{F1@\{10, 25, 50\}}} \\
        \hline
        & TCN \cite{lea2017temporal} & 88.3 & - & - & - & 56.8 & 61.5 & - & - & - & 20.0 \\
        Supervised & ST-GCN \cite{yan2018spatial} & 88.7 & - & - & - & 67.7 & 67.9 & - & - & - & 25.8 \\
         & MS-TCN \cite{farha2019ms} & 86.8 & - & - & - & 89.9 & 65.8 & - & - & - & 39.6 \\
        & MS-GCN \cite{filtjens2022skeleton} & 90.4 & - & - & - & 93.0 & 65.6 & - & - & - & 43.6 \\
        \hline
        & CTE \cite{kukleva2019unsupervised} & 33.8 & 4.7 & 0.6 & 0.6 & 0.5 & 23.3 & 16.8 & 8.1 & 5.2 & 2.3 \\
        & CTE + Viterbi \cite{kukleva2019unsupervised} & 39.2 & 21.7 & 13.2 & 9.5 & 7.5 & 23.0 & 17.7 & 6.8 & 3.7 & 1.6 \\
        Unsupervised & TOT \cite{kumar2022unsupervised} & 33.8 & 3.1 & 0.7 & 0.5 & 0.4 & 21.4 & 7.8 & 5.3 & 2.7 & 1.1 \\
        & TOT + Viterbi \cite{kumar2022unsupervised}  & 33.8 & 20.8 & 15.6 & 10.5 & 7.5 & 32.6 & 17.7 & 11.6 & 7.4 & 3.2 \\
        & ASOT \cite{xu2024temporally} & 33.9 & 17.4 & 4.5 & 3.8 & 3.0 & 22.9 & 23.4 & 17.8 & 12.1 & 5.7 \\
         & \textbf{\model\ (ours)} & \textbf{42.0} & \textbf{36.1} & \textbf{38.5} & \textbf{31.5} & \textbf{24.3} & 
        \textbf{37.4} & \textbf{39.4} & \textbf{34.7} & \textbf{28.4} & \textbf{16.4} \\
        \hline
    \end{tabular}
    \caption{Comparison to supervised and unsupervised temporal action segmentation methods on the HuGaDB and LARa datasets.\label{tab:tas_comparison-hugadblara}}
    \vspace{-2mm}
\end{table*}

\begin{table*}[t]
    \centering
    \resizebox{\textwidth}{!}{
    \begin{tabular}{@{}l|lllll|lllll|lllll@{}}
        \hline
        & \multicolumn{5}{c|}{\textbf{BABEL Subset-1}} & \multicolumn{5}{c|}{\textbf{BABEL Subset-2}} & \multicolumn{5}{c}{\textbf{BABEL Subset-3}} \\
        \cline{2-16}
        \textbf{Method} & \textbf{MoF} & \textbf{Edit} & \multicolumn{3}{c|}{\textbf{F1@\{10, 25, 50\}}} & \textbf{MoF} & \textbf{Edit} & \multicolumn{3}{c|}{\textbf{F1@\{10, 25, 50\}}} & \textbf{MoF} & \textbf{Edit} & \multicolumn{3}{c}{\textbf{F1@\{10, 25, 50\}}} \\
        \hline
        CTE \cite{kukleva2019unsupervised} & 34.8 & 28.6 & 25.0 & 17.5 & 9.5 &
        40.3 & 30.6 & 17.8 & 12.2 & 7.4 &
        31.4 & 13.1 & 8.2 & 5.8 & 3.6 \\
        
        CTE + Viterbi \cite{kukleva2019unsupervised} & 30.9 & 36.2 & 23.2 & 15.2 & 7.3 
        & 42.4 & 30.7 & 24.3 & 19.5 & 12.8
        & 31.2 & 30.9 & 20.7 & 15.2 & 8.4 \\
        
        TOT \cite{kumar2022unsupervised} & 31.8 & 18.7 & 14.2 & 7.6 & 4.4 & 
        35.4 & 12.8 & 13.7 & 8.6 & 4.3 & 
        31.5 & 7.1 & 4.9 & 2.9 & 1.7 \\
        
        TOT + Viterbi \cite{kumar2022unsupervised}  & 29.1 & 29.3 & 31.5 & 20.8 & 9.9 &
        35.3 & 36.8 & 35.9 & 30.0 & 19.8 &
        34.0 & 33.8 & 31.3 & 26.8 & 17.9\\
        
        ASOT \cite{xu2024temporally} & 35.3 & \textbf{43.1} & \textbf{42.3} & \textbf{34.1} & \textbf{24.5} &
        43.1 & 37.7 & 40.3 & 33.4 & 23.4 &
        38.0 & 27.1 & 27.4 & 21.6 & 14.3 \\

        \textbf{\model\ (ours)} & \textbf{36.6} & 38.5 & 40.9 & 32.8 & 22.3 &
        \textbf{49.1} & \textbf{37.8} & \textbf{43.8} & \textbf{37.4} & \textbf{27.4} &
        \textbf{40.6} & \textbf{38.6} & \textbf{38.0} & \textbf{29.3} & \textbf{19.3} \\
        \hline
    \end{tabular}}
    \caption{Comparison to unsupervised temporal action segmentation methods on the BABEL dataset.\label{tab:tas_comparison-babel}}
\end{table*}

\subsection{Datasets}
We evaluated our method on three skeleton-based action datasets: HuGaDB \cite{chereshnev2018hugadb}, LARa \cite{niemann2020lara}, and BABEL \cite{BABEL:CVPR:2021}. Additional results on the PKU-MMD v2 dataset~\cite{liu2017pku} are reported in the suppl.\ material. 

\textbf{HuGaDB} consists of 364 trials capturing 10 lower limb activities, including walking, running, and sitting. Data were collected from 18 participants using six IMUs (3-axis accelerometer and 3-axis gyroscope) placed on the thighs, shins, and feet, recorded at 60 Hz.

\textbf{LARa} comprises 439 trials across 8 warehouse-related action categories. Data were collected from 16 participants using marker-based MoCap (200 fps), recording the 3-axis position and orientation of 22 full body joints. The dataset includes 960 minutes of recordings.
LARa is downsampled by a factor of 4, reducing the frame rate to 50 fps \cite{filtjens2022skeleton}. Additionally, the skeleton is centered from the root joint for translation invariance.

The \textbf{BABEL} dataset contains 43 hours of 3D human motion sequences from AMASS \cite{AMASS:ICCV:2019}, annotated with over 63,000 frame-level labels across more than 250 action categories. Following the protocol of \cite{yu2023frame}, we extract 25 full-body joint skeletons from SMPL~\cite{SMPL:2015} parameters,  construct three subsets, each focusing on four action categories, and downsample all sequences to 30 fps, after centering them on the root joint. In addition to the original protocol, we exclude sequences containing more than 50\% background actions to ensure that the models learn relevant actions. The evaluation metrics and implementation details are described in the suppl.\ material.

\subsection{Comparison to State of the Art}
We compare our proposed method, \textit{\model}, with state-of-the-art unsupervised temporal action segmentation methods on the HuGaDB, LARa and BABEL datasets, as shown in Tables~\ref{tab:tas_comparison-hugadblara} and \ref{tab:tas_comparison-babel}. Across all evaluation metrics, \textit{\model} consistently outperforms unsupervised temporal action segmentation approaches.

On the HuGaDB dataset in Table~\ref{tab:tas_comparison-hugadblara}, \textit{\model} achieves a MoF of 42.0 and Edit Score of 36.1, surpassing the best performing approach CTE~\cite{kukleva2019unsupervised} by 2.8 and 14.4, respectively.   
For the F1@50 metric, our method has the score of 24.3, which is by 16.8 higher than CTE \cite{kukleva2019unsupervised} and TOT \cite{kumar2022unsupervised}. 
On the LARa dataset, \textit{\model} achieves a MoF of 37.4, Edit Score of 39.4, and F1@50 of 16.4, outperforming  the best approach ASOT \cite{xu2024temporally} by a large margin as well. For completeness, we also include results from supervised approaches, which reported only MoF and F1@50. While supervised approaches achieve much higher scores, unsupervised approaches do not use any annotations and solve thus a much harder task.        

\begin{figure*}[t]
    \centering
    \begin{subcaptionbox}{HuGaDB\label{fig:hugadb}}{
        \includegraphics[draft=false,width=0.48\textwidth]{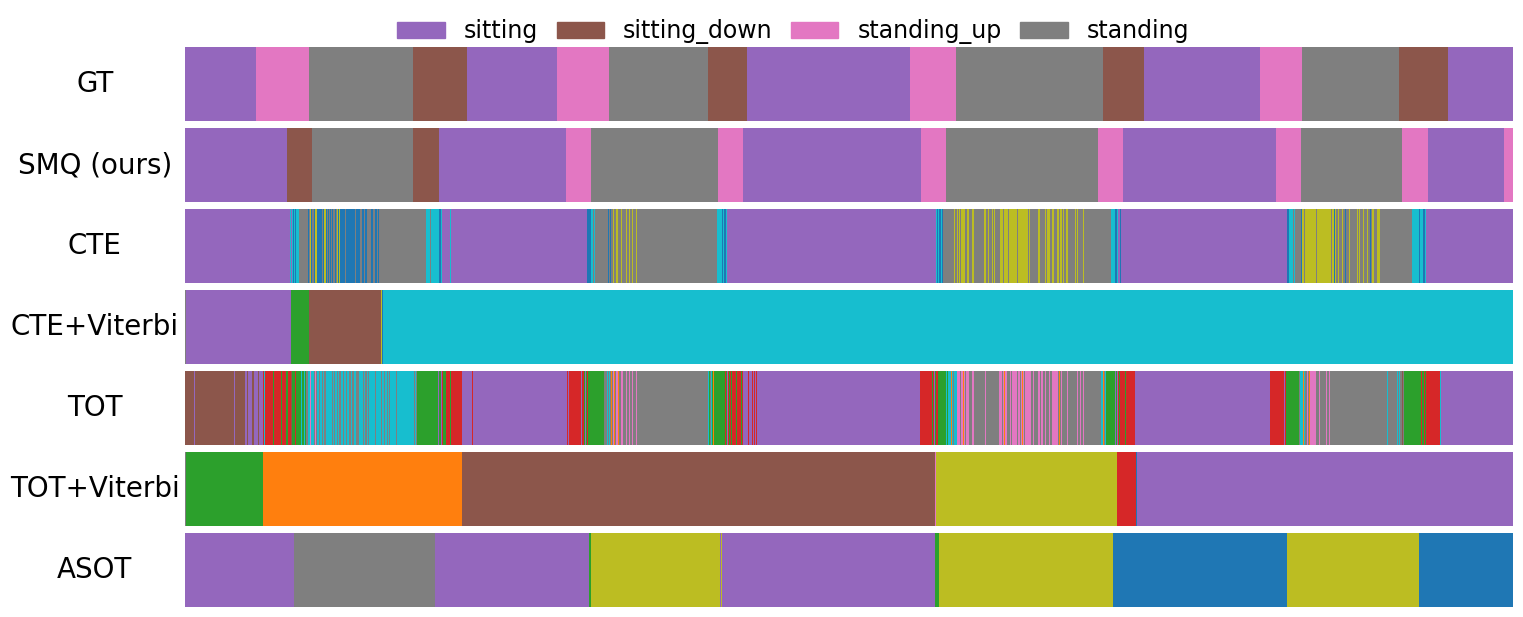}
    }\end{subcaptionbox}
    \begin{subcaptionbox}{LARa\label{fig:lara}}{
        \includegraphics[width=0.48\textwidth]{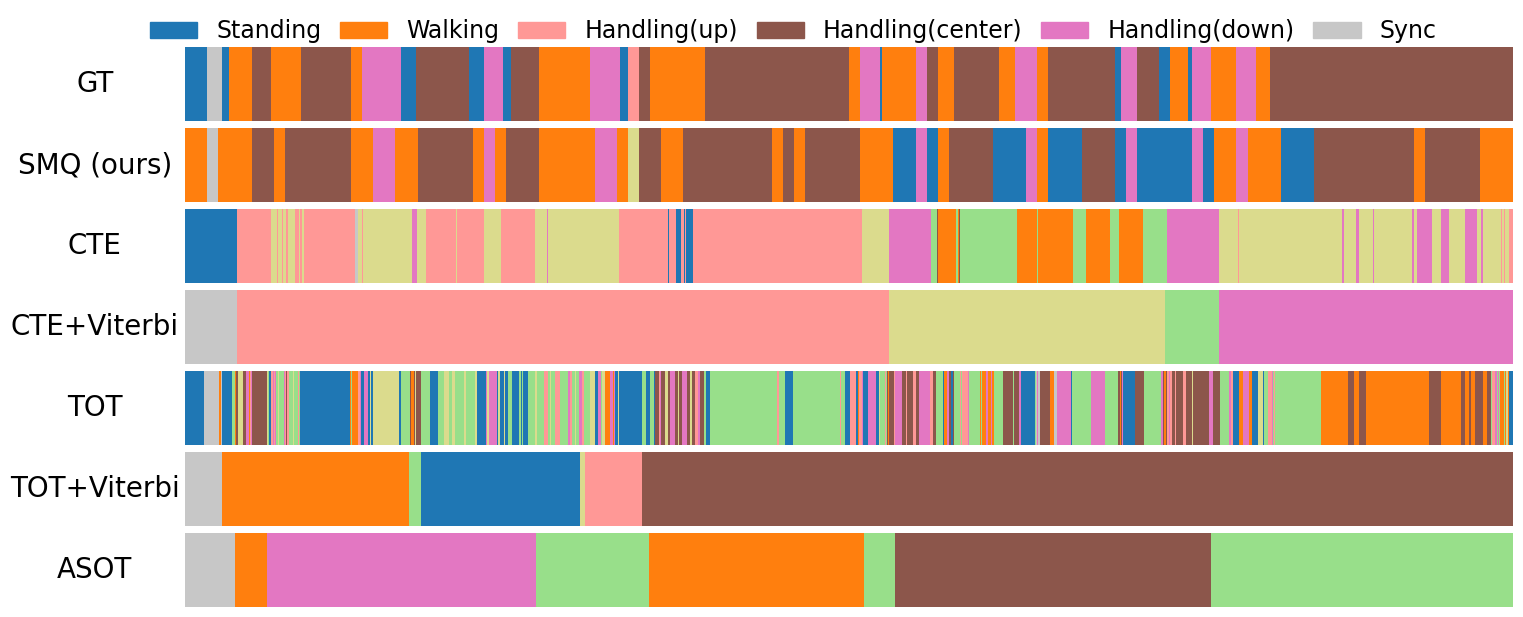}
    }\end{subcaptionbox}
    \begin{subcaptionbox}{BABEL Subset-1\label{fig:babel1}}{
        \includegraphics[width=0.48\textwidth]{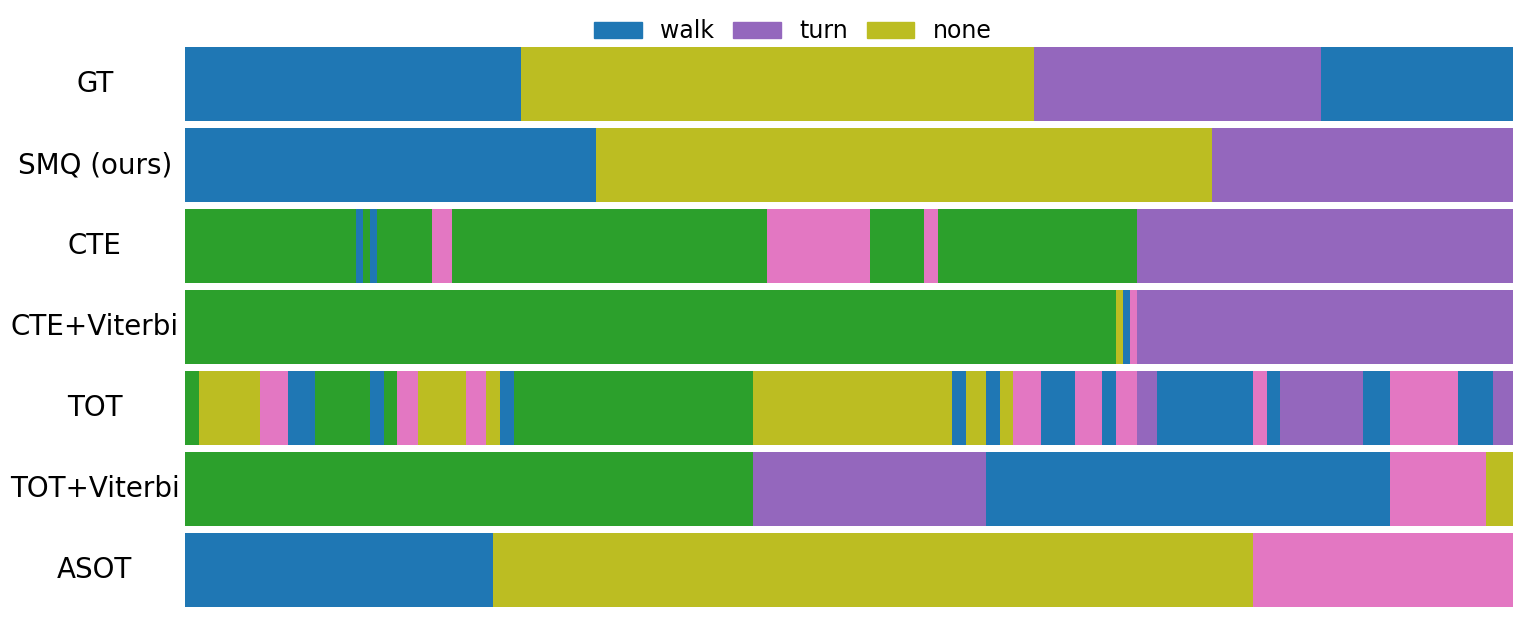}
    }\end{subcaptionbox}
    \begin{subcaptionbox}{BABEL Subset-3\label{fig:babel3}}{
        \includegraphics[width=0.48\textwidth]{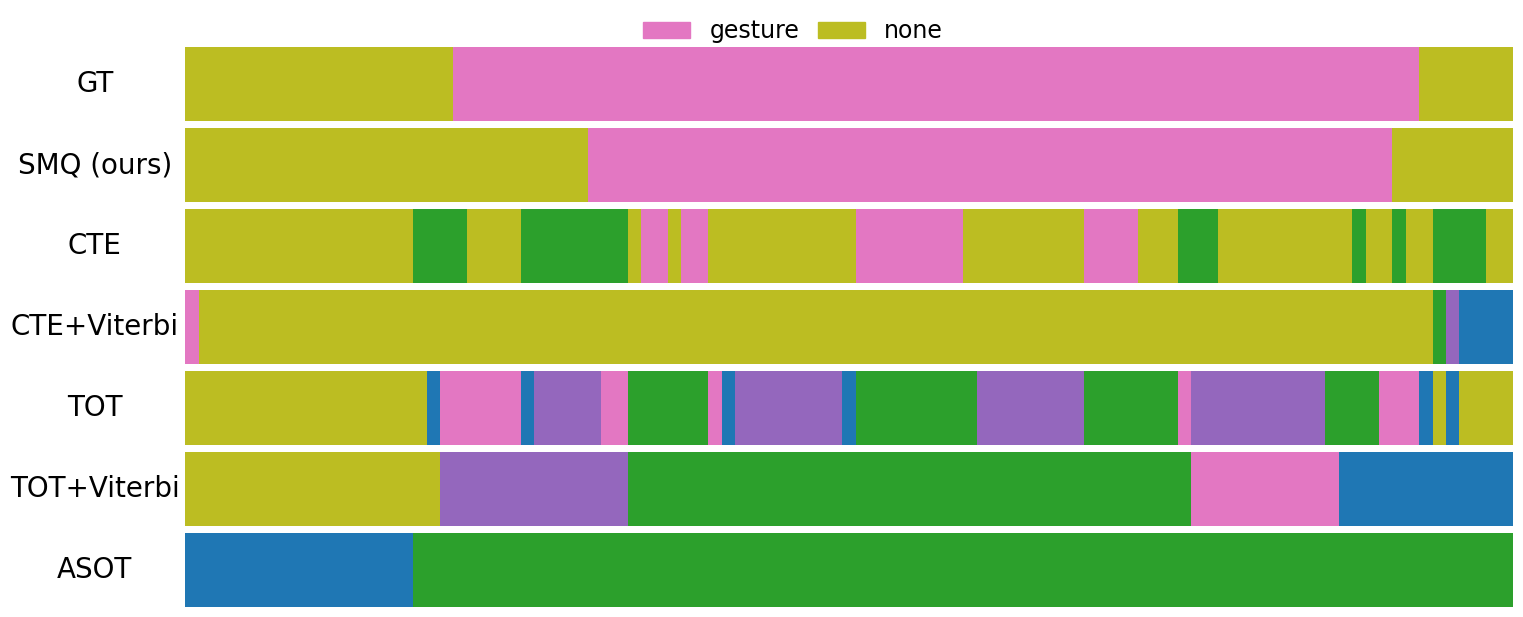}
    }\end{subcaptionbox}
    \vspace{-2mm}
    \caption{Qualitative results for unsupervised action segmentation algorithms.\label{fig:three_figures}}
    \vspace{-2mm}
\end{figure*}

On the BABEL dataset in Table~\ref{tab:tas_comparison-babel}, \textit{\model} achieves a higher MoF than ASOT~\cite{xu2024temporally} for Subset-1, but ASOT performs slightly better for the other metrics. Whereas on Subset-2 and Subset-3, \textit{\model} outperforms all unsupervised temporal action segmentation methods. 

Fig.~\ref{fig:three_figures} shows a qualitative comparison between the methods. 
Both CTE and TOT generate too many small segments. They thus employ a post-processing technique utilizing Viterbi decoding \cite{8585084,ding2023temporal}, to refine initial predictions and achieve temporally consistent action segments. The Viterbi decoding \cite{8585084} computes the most probable sequence of actions by maximizing the joint likelihood under a first-order Markov model, using transition probabilities that penalize frequent label changes. 
While this improves segmental metrics by reducing fragmentation, it also imposes a strong prior that each action appears only once per sequence, which is not the case in the datasets.

ASOT can identify re-occurring actions and does not suffer from small segments. The approach, however, has the tendency to predict rather large segments. 
The small segments in the HuGaDB and LARa datasets are not discovered. Since the segments are longer in the BABEL subsets, ASOT performs better on this dataset. While \textit{\model} does not generate perfect results, for instance it mixes standing up and sitting down in the example from the HuGaDB dataset, the results are better aligned with the ground truth. It is able to detect re-occurring actions, and it does not have the tendency to generate too many small segments, as CTE and TOT, or overly large segments as ASOT. Overall, the results show that \textit{\model} is more effective in handling diverse segmentation scenarios. However, as can be seen in Fig.~\ref{fig:three_figures}, identifying precise action transitions remains a challenge. Since actions are assigned at the patch level, transitions can only occur at patch boundaries. Improving the detection of action boundaries needs to be investigated in future work.

\begin{table*}[t]
    \centering
    \resizebox{\textwidth}{!}{
    \begin{tabular}{@{}l|lllll|lllll|lllll@{}}
        \hline
        & \multicolumn{5}{c|}{\textbf{HuGaDB}} & \multicolumn{5}{c|}{\textbf{LARa}} & \multicolumn{5}{c}{\textbf{BABEL - Avg}} \\
        \cline{2-16}
        \textbf{Method} & \textbf{MoF} & \textbf{Edit} & \multicolumn{3}{c|}{\textbf{F1@\{10, 25, 50\}}} & \textbf{MoF} & \textbf{Edit} & \multicolumn{3}{c|}{\textbf{F1@\{10, 25, 50\}}} & \textbf{MoF} & \textbf{Edit} & \multicolumn{3}{c}{\textbf{F1@\{10, 25, 50\}}} \\
        \hline
        Predict\&Cluster \cite{su2020predict} & 39.3 & 31.5 & 33.2 & 24.2 & 18.3 &
        18.8 & 28.5 & 19.2 & 12.9 & 5.4 &
        32.2 & 32.1 & 32.2 & 24.3 & 14.2 \\
        
        AimCLR \cite{guo2022contrastive} & 34.7 & 30.6 & 30.9 & 23.7 & 11.3 
        & 22.0 & 28.0 & 22.1 & 15.6 & 7.7
        & 40.3 & 29.7 & 32.5 & 25.7 & 15.5 \\
        
        ActCLR \cite{lin2023actionlet} & 28.2 & 21.6 & 10.4 & 6.2 & 4.1 & 
        23.3 & 30.6 & 23.7 & 16.7 & 7.8 & 
        32.6 & 33.2 & 32.8 & 24.4 & 13.4 \\
        
        \textbf{\model\ (ours)} & \textbf{42.0} & \textbf{36.1} & \textbf{38.5} & \textbf{31.5} & \textbf{24.3} & \textbf{37.4} & \textbf{39.4} & \textbf{34.7} & \textbf{28.4} & \textbf{16.4} & \textbf{42.1} & \textbf{38.3} & \textbf{40.9} & \textbf{33.2} & \textbf{23.0} \\
        \hline
    \end{tabular}}
    \caption{Comparison to self-supervised skeleton-based representation learning methods.\label{tab:sar_comparison}}
    \vspace{-4mm}
\end{table*}

We also compare \textit{\model} to self-supervised representation learning approaches for skeleton data~\cite{su2020predict,guo2022contrastive,lin2023actionlet}. These approaches are typically first trained in a self-supervised way and then fine-tuned for different downstream tasks like skeleton-based action recognition using annotated data. In order to apply these methods to unsupervised temporal action segmentation, we divided the input skeleton sequences into non-overlapping patches, consistent with the latent patching in \textit{\model}. These patches were fed through the self-supervised representation learning approaches to learn patch-level action representations. We then performed k-means clustering over the patch-level representations. 
The results in Table~\ref{tab:sar_comparison} show that \textit{\model} consistently outperforms existing self-supervised representation learning approaches on all the datasets. It needs to be noted that these approaches are designed for datasets with only one action per sequence, whereas unsupervised temporal action segmentation requires processing long, untrimmed skeleton sequences containing multiple actions per sequence. Compared to \textit{\model}, they therefore learn less distinctive representations for this task as it is visualized in the suppl.\ material.

\begin{table*}
    \centering
    \resizebox{\textwidth}{!}{
    \begin{tabular}{@{}l|lllll|lllll|lllll@{}}
        \hline
         & \multicolumn{5}{c|}{\textbf{HuGaDB}} & \multicolumn{5}{c|}{\textbf{LARa}} & \multicolumn{5}{c}{\textbf{BABEL - Avg}} \\
        \cline{2-16}
        \textbf{Embedding} & \textbf{MoF} & \textbf{Edit} & \multicolumn{3}{c|}{\textbf{F1@\{10, 25, 50\}}} & \textbf{MoF} & \textbf{Edit} & \multicolumn{3}{c|}{\textbf{F1@\{10, 25, 50\}}} & \textbf{MoF} & \textbf{Edit} & \multicolumn{3}{c}{\textbf{F1@\{10, 25, 50\}}} \\
        \hline
        Entangled & 38.9 & 34.2 & 35.6 & 27.9 & 18.9  &
        26.6 & 35.9 & 29.5 & 22.8 & 12.3 &
        40.1 & 32.1 & 34.9 & 27.1 & 16.9 \\
        \textbf{Disentangled} & \textbf{42.0} & \textbf{36.1} & \textbf{38.5} & \textbf{31.5} & \textbf{24.3} & 
        \textbf{37.4} & \textbf{39.4} & \textbf{34.7} & \textbf{28.4} & \textbf{16.4}
        & \textbf{42.1} & \textbf{38.3} & \textbf{40.9} & \textbf{33.2} & \textbf{23.0} \\
        \hline
    \end{tabular}}
    \caption{Impact of disentangled joint embedding on the HuGaDB, LARa, and BABEL datasets.\label{tab:joint_independence}}
    \vspace{-5mm}
\end{table*}

\subsection{Ablations}\label{sec:ablation}
In this section, we analyze the impact of different design choices in our model.

\noindent\textbf{Impact of Disentangled Embedding.}\label{sec:ablation-channel-independence}
As illustrated in Fig.~\ref{fig:smq-teaser}, the encoder processes each joint independently, which ensures that the information of the different joints remains disentangled in the embedding space. To evaluate the effectiveness of the disentangled embedding space, we conducted an ablation study. In Table~\ref{tab:joint_independence}, we compare our approach to a setup where we concatenate the joints before learning an embedding. In the latter case, the joints get entangled in the embedding and the structural information of the skeleton gets lost. The results demonstrate that the disentangled embedding enhances our model's performance, indicating that learning joint representations separately allows for more accurate action segmentation by enabling each joint to reside in distinct subspaces.

\begin{table}[t]
  \centering
  \setlength{\tabcolsep}{4pt}
  \begin{tabular}{@{}llllll@{}}
    \toprule
    \textbf{Patch Size} &
    \textbf{MoF} &
    \textbf{Edit} &
    \multicolumn{3}{c}{\textbf{F1@\{10, 25, 50\}}} \\
    \midrule
    1 (Frame-wise) & 33.9 & 25.6 & 21.2 & 14.9 & 8.6 \\
    25 (0.5 sec) & 35.2 & 35.7 & 32.4 & 25.0 & 14.9 \\
    
    \textbf{50 (1 sec)} & \textbf{37.4} & \textbf{39.4} & \underline{\textbf{34.7}} & \underline{\textbf{28.4}} & 16.4\\

    75 (1.5 sec) & 37.2 & 39.1 & \textbf{\underline{34.7}} & \underline{\textbf{28.4}} & \textbf{17.1}\\
    
    100 (2 sec) & 30.8 & 36.1 & 33.0 & 27.0 & 15.4 \\
    
    150 (3 sec) & 31.4 & 34.0 & 32.6 & 26.5 & 14.7 \\
    \bottomrule
  \end{tabular}
  \caption{Impact of motion patch size on the LARa dataset.\label{tab:patch-size}}
  \vspace{-2mm}
\end{table}

\noindent\textbf{Impact of Patch Size.}\label{sec:ablation-patch-size}
We apply temporal patching to the latent skeleton embeddings to obtain skeleton motion words.
Specifically, we use a fixed patch size to create non-overlapping patches of the representations. 
To investigate the effect of patch size on the performance of our model, we conducted an ablation study with varying patch sizes, as presented in Table~\ref{tab:patch-size}. Without patching, \ie, patch size is 1, we obtain a frame-wise representation. The results indicate that \textit{\model} consistently outperforms the frame-wise representation across all evaluated metrics, with significant improvements in segmental metrics. These findings emphasize that frame-wise quantization leads to over-segmentation, as it fails to capture the inherent temporal consistency of actions. In contrast, \textit{\model} captures the temporal variability of actions more effectively by grouping contiguous latent embeddings into patches. This results in more semantically meaningful action clusters that yield temporally consistent segmentations. On the LARa dataset, a patch size of 50 performs best, which corresponds to one second of motion, as the data is downsampled to 50 fps. We keep this one-second patching consistent across all datasets.

\noindent\textbf{Impact of Loss Variants.}\label{sec:ablation-loss}
To evaluate the effectiveness of different loss components, we conduct an ablation study examining various configurations of the commitment loss and reconstruction loss. We first evaluate three variants of the reconstruction loss in the last three rows of Table~\ref{tab:ablation-loss}. The proposed inter-joint distance MSE loss performs better than the commonly used MSE loss since it is invariant to translation and rotation. The root distance MSE loss is obtained by computing the MSE over the distances of each joint to the root joint within the same frame. While this loss is also translation and rotation invariant, it performs even worse than the inter-joint variant. Furthermore, we evaluate the impact of the commitment loss and reconstruction loss by omitting one of them as presented in the first two rows of the Table~\ref{tab:ablation-loss}. Removing either of the loss terms substantially decreases the performance.

\begin{table}[tb]
  \resizebox{\columnwidth}{!}{
  \centering
  \setlength{\tabcolsep}{4pt}
  \begin{tabular}{@{}c|c|lllll@{}}
    \toprule
    \textbf{$L_{commit}$} &
    \textbf{Reconstruction Loss} &
    \textbf{MoF} &
    \textbf{Edit} &
    \multicolumn{3}{c}{\textbf{F1@\{10, 25, 50\}}} \\
    \midrule
    \checkmark & \ding{55} & 29.9 & 29.9 & 26.8 & 17.7 & 8.8 \\
    
    \ding{55} & Inter-Joint Dist. MSE & 31.0 & 34.6 & 30.6 & 24.4 & 14.4 \\
    \midrule
     \checkmark & MSE & 34.6 & \underline{\textbf{39.4}} & 34.5 & 27.4 & 15.7 \\
    
    \checkmark & Root Distance MSE & 28.0 & 34.9 & 30.2 & 23.8 & 14.1  \\

    \checkmark & \textbf{Inter-Joint Dist. MSE} & \textbf{37.4} & \underline{\textbf{39.4}} & \textbf{34.7} & \textbf{28.4} & \textbf{16.4} \\
    \bottomrule
  \end{tabular}}
  \caption{Impact of loss function variants on the LARa dataset.\label{tab:ablation-loss}}
  \vspace{-2mm}
\end{table}

\noindent\textbf{Impact of Reconstruction Loss Weight.}\label{sec:ablation-loss-weight}
In our ablation study, we varied the reconstruction loss weight, $\lambda$~\eqref{eq:total-loss}, to understand its effect on model performance. By adjusting $\lambda$, we explore the balance between the reconstruction and commitment loss in Table~\ref{tab:rec-weight}. Our model is not very sensitive to this parameter and performs well for $\lambda=0.001$.

\begin{table}[t]
  \centering
  \setlength{\tabcolsep}{4pt} 
  \begin{tabular}{@{}llllll@{}}
    \toprule
    \textbf{$\lambda$} &
    \textbf{MoF} &
    \textbf{Edit} &
    \multicolumn{3}{c}{\textbf{F1@\{10, 25, 50\}}} \\
    \midrule
    0.1 & 35.9 & 38.4 & 33.4 & 26.9 & 15.6 \\

    0.01 & 36.2 & 38.9 & 34.4 & 27.8 & 15.9 \\

    \textbf{0.001} & \textbf{37.4} & 39.4 & 34.7 &
    \textbf{28.4}& \textbf{16.4} \\
    
    0.0001 & 34.1 & \textbf{39.8} & \textbf{35.4} & 27.9 & 16.1 \\
    \bottomrule
  \end{tabular}
  \caption{Impact of $\lambda$ on the LARa dataset.\label{tab:rec-weight}}
  \vspace{-2mm}
\end{table}

\noindent\textbf{Impact of Decay Factor.}\label{sec:ablation-decay}
To examine the role of the EMA decay factor $\alpha$~\eqref{eq:ema_update}, we perform an ablation study by evaluating multiple $\alpha$ values in Table~\ref{tab:decay-rate}. The decay factor governs the balance between incorporating new patch information and maintaining historical motion word estimates, thereby influencing the smoothness and responsiveness of the updates. Our model performs best when $\alpha$ is set to 0.5. Additional ablation studies are provided in the suppl.\ material.

\begin{table}[tb]
  \centering
  \setlength{\tabcolsep}{4pt} 
  \begin{tabular}{@{}llllll@{}}
    \toprule
    \textbf{$\alpha$} &
    \textbf{MoF} &
    \textbf{Edit} &
    \multicolumn{3}{c}{\textbf{F1@\{10, 25, 50\}}} \\
    \midrule
    0.35 & 34.8 & 39.4 & \textbf{35.0} & 27.6 & 15.5 \\

    \textbf{0.50} & \textbf{37.4} & \textbf{39.4} & 34.7 &
    \textbf{28.4}& \textbf{16.4} \\

    0.65 & 35.5 & 38.5 & 33.9 & 27.2 & 15.8 \\
    0.80 & 34.2 & 37.7 & 33.4 & 27.1 & 15.5 \\
    \bottomrule
  \end{tabular}
  \caption{Impact of EMA decay rate ($\alpha$) on the LARa dataset.\label{tab:decay-rate}}
  \vspace{-2mm}
\end{table}
\section{Conclusion}

In this paper, we introduced \textit{\model}, a method for unsupervised skeleton-based temporal action segmentation. It uses an autoencoder that preserves the skeleton structure by ensuring that the information of the different joints remains disentangled in the embedding space. The skeleton motion quantization module then patches the latent representations into short time series of fixed length and quantizes them into skeleton motion words. We performed an extensive evaluation and compared the proposed approach to state-of-the-art unsupervised temporal action segmentation approaches, as well as approaches for self-supervised skeleton-based representation learning. \textit{\model} outperforms them by a large margin. Furthermore, we demonstrated the importance of learning the representation over patches instead of frames and the importance of an embedding space where the information of the different joints remains disentangled. As part of future work, it would be interesting to investigate a hierarchical version of \textit{\model} and improve the detection of action boundaries.  

\section*{Acknowledgments}
This work has been supported by the ERC Consolidator Grants FORHUE (101044724) and ActionContraThreat (816564), the Ministry of Culture and Science of the State of North Rhine-Westphalia (iBehave network, Programme ``Netzwerke 2021"), and the TRA Modelling (University of Bonn) as part of the Excellence Strategy of the federal and state governments.

{
    \small
    \bibliographystyle{ieeenat_fullname}
    \bibliography{main}
}
\clearpage
\clearpage
\setcounter{page}{1}
\maketitlesupplementary

\section{Implementation Details and Evaluation Metrics} 

\subsection{Implementation Details}
In \textit{\model}, both the encoder and decoder use a two-stage Temporal Convolutional Network (MS-TCN), with each stage comprising three dilated residual layers to effectively capture temporal dependencies. The codebook size corresponds to the number of ground-truth actions in the dataset as it is required by the protocol. The patch size is fixed to cover one second of frames, adjusted according to each dataset’s fps. For the codebook updates, we use an exponential moving average (EMA) with a decay factor of 0.5, and $\lambda$ is set to 0.001. The model is trained using the Adam optimizer with a learning rate of 0.0005, with a batch size of 8 for the LARa and HuGaDB datasets and 32 for the BABEL subsets. Training is conducted on a single NVIDIA RTX 4090 GPU.

\subsection{Evaluation Metrics}
In the unsupervised temporal action segmentation setting, the predicted clusters from the model do not inherently correspond to the ground truth actions. To address this, we employ the global Hungarian matching algorithm following the previous methods \cite{kukleva2019unsupervised, kumar2022unsupervised, xu2024temporally, sener2018unsupervised, li2021action, kumar2022unsupervised}, which establishes a one-to-one mapping between predicted segments and ground truth labels across the entire dataset. This mapping is used for calculating evaluation metrics, ensuring that each predicted cluster is properly aligned with its corresponding ground truth action.

We report both frame-based and segment-based metrics \cite{ding2023temporal, farha2019ms, LaSA_2024ECCV}. Mean over frames (MoF) measures the proportion of correctly predicted frames but does not account for over-segmentation. To better assess prediction quality, we also report segmental metrics: the edit score \cite{lea2016segmental}, based on the Levenshtein distance, and the segmental F1 score \cite{lea2017temporal} at Intersection over Union (IoU) thresholds of 10\%, 25\%, and 50\% (F1@{10, 25, 50}). These metrics provide a more comprehensive evaluation by penalizing over-segmentation and capturing alignment between predicted and ground truth segments.

\section{Results on PKU-MMD v2}
PKU-MMD v2~\cite{liu2017pku} contains 1009 skeleton sequences spanning 41 action categories, performed by 13 subjects. Each sequence lasts approximately 1 to 2 minutes and includes around 7 action instances. The data were recorded at 30 fps using a Kinect v2 sensor. Each frame provides the 3D positions of 25 full body joints. To prepare the dataset, we centered the skeletons from the root joint to ensure translation invariance.

\begin{figure}[t]
    \centering
    \begin{subfigure}{0.5\textwidth}
        \centering
        \includegraphics[width=\textwidth]{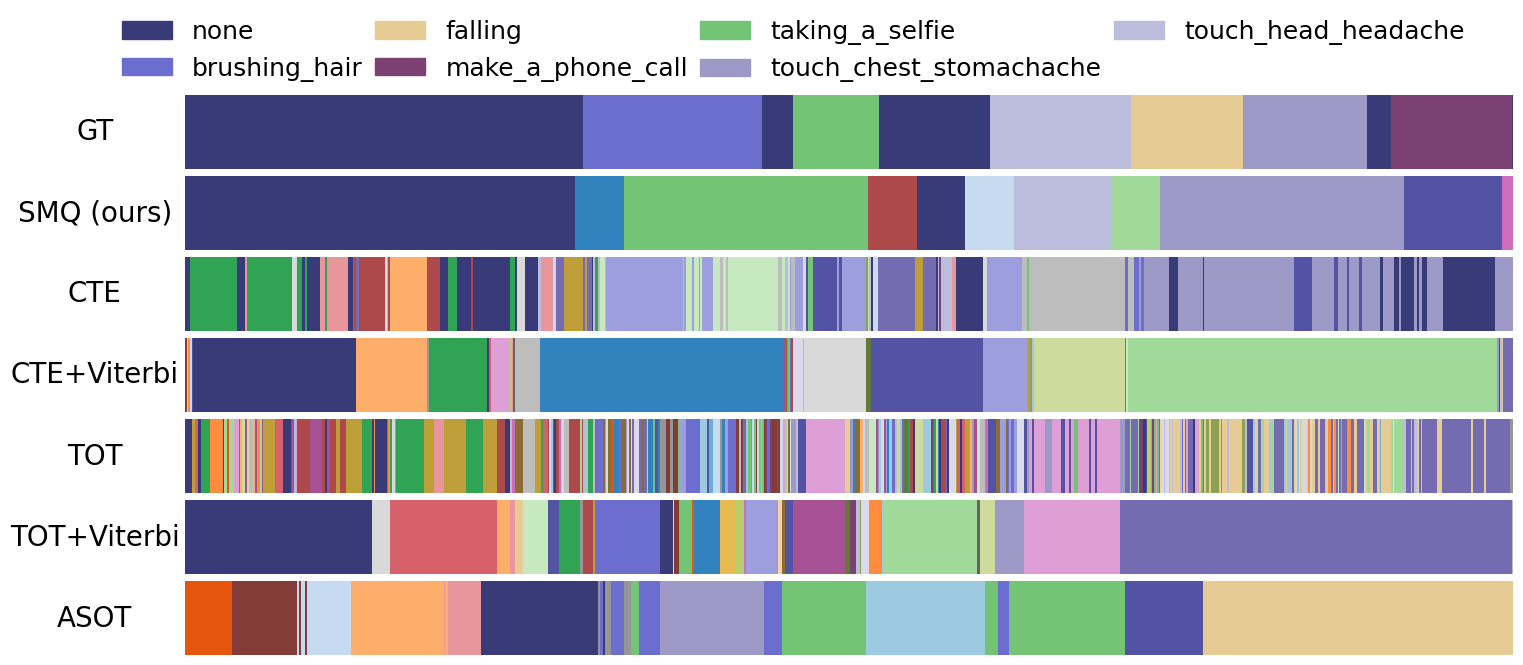}
        \label{fig:subfigure1-pku}
    \end{subfigure}
    \begin{subfigure}{0.5\textwidth}
        \centering
        \includegraphics[width=\textwidth]{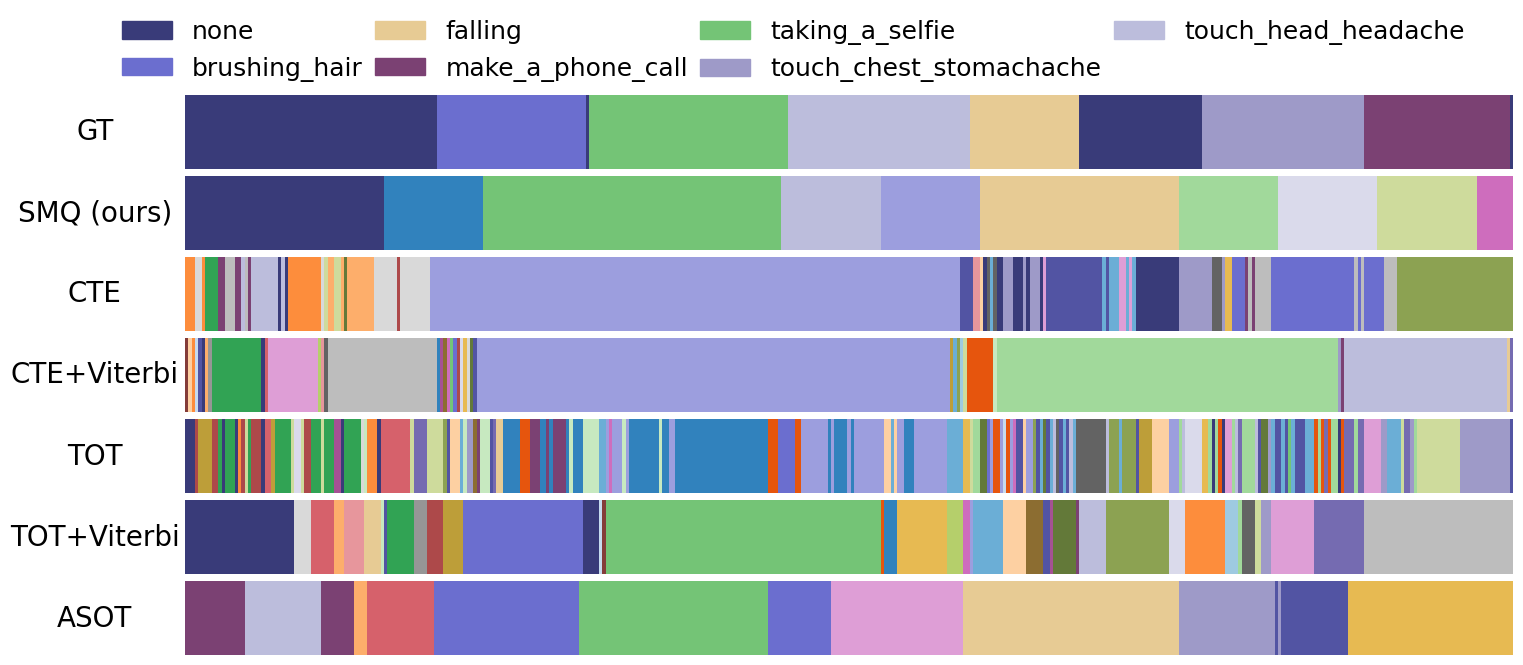}
        \label{fig:subfigure2-pku}
    \end{subfigure}
    \begin{subfigure}{0.5\textwidth}
        \centering
        \includegraphics[width=\textwidth]{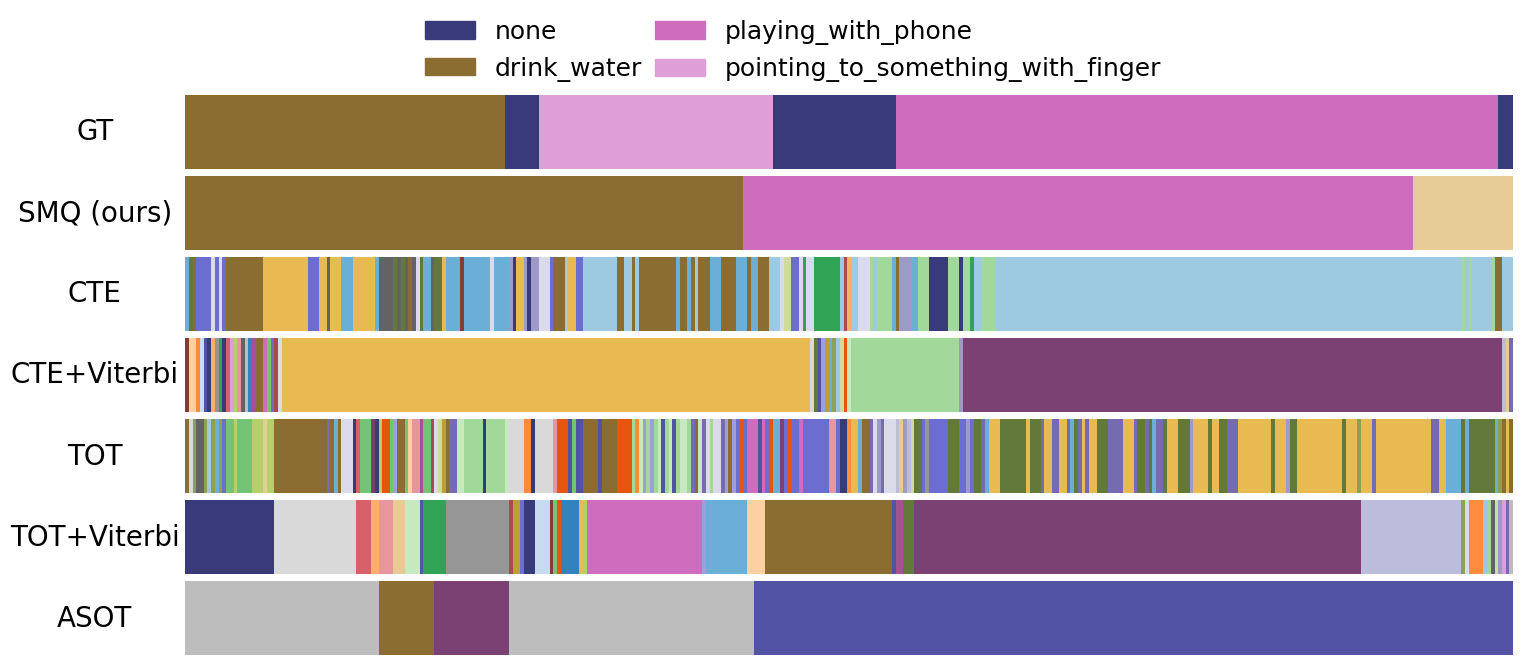}
        \label{fig:subfigure3-pku}
    \end{subfigure}
    \caption{Qualitative results for unsupervised action segmentation algorithms on the PKU-MMD v2 dataset.}
    \label{fig:qualitative-pkummd}
\end{figure}

As shown in Table~\ref{tab:results-pkummd}, \textit{SMQ} achieves the best performance across all metrics except for MoF, where CTE + Viterbi is slightly better. The overall scores remain low due to the challenging nature of the PKU-MMD v2 dataset, which includes 41 action categories and approximately 40\% background (none) frames, making it particularly difficult for unsupervised temporal action segmentation. We also show qualitative results on PKU-MMD v2 in Figure~\ref{fig:qualitative-pkummd}.

\begin{table}[t]
  \centering
  \setlength{\tabcolsep}{4pt}
  \begin{tabular}{@{}l|lllll@{}}
    \toprule
    \textbf{Method} &
    \textbf{MoF} &
    \textbf{Edit} &
    \multicolumn{3}{c}{\textbf{F1@\{10, 25, 50\}}} \\
    \midrule
    CTE~\cite{kukleva2019unsupervised} & 8.6 & 4.5 & 1.8 & 1.0 & 0.4 \\
    CTE + Viterbi~\cite{kukleva2019unsupervised} & 8.1 & 10.8 & 3.4 & 2.3 & 1.0 \\
    TOT~\cite{kumar2022unsupervised} & 6.6 & 3.0 & 0.6 & 0.2 & 0.1 \\
    TOT + Viterbi~\cite{kumar2022unsupervised} & \textbf{15.1} & 10.8 & 5.8 & 4.2 & 2.2\\
    ASOT~\cite{xu2024temporally} & 9.0 & 9.4 & 6.0 & 4.4 & 2.4 \\
    \midrule
    \textbf{SMQ (ours)} & 13.2 & \textbf{13.8} & \textbf{13.8} & \textbf{10.6} & \textbf{5.6} \\
    \bottomrule
  \end{tabular}
  \caption{Comparison to unsupervised temporal action segmentation methods on the PKU-MMD v2 dataset.\label{tab:results-pkummd}}
\end{table}

\section{Additional Ablations}

\subsection{Impact of Disentangled Embedding}\label{sec:channel-independence-all}

To further analyze the impact of the disentangled embedding, we conducted an ablation study comparing two asymmetric encoder-decoder configurations. In the first configuration, the encoder processes each joint independently, while the decoder concatenates the features to reconstruct the skeleton. In the second configuration, the encoder concatenates all joint features, which are then processed separately in the decoder. Results in Table~\ref{tab:asymmetric} show that independently processing joints in both the encoder and decoder yields the best performance, demonstrating the critical role of keeping joints disentangled throughout the architecture. Moreover, processing joints independently in the encoder alone achieves the second-best performance, highlighting the significance of encoding joint-specific features for capturing fine-grained skeleton dynamics, which appears fundamental for effective skeleton-based temporal action segmentation.

\begin{table}
  \resizebox{\columnwidth}{!}{
  \centering
  \setlength{\tabcolsep}{6pt} 
  \begin{tabular}{@{}cc|ccccc@{}}
    \toprule
    \multicolumn{2}{c|}{\textbf{Independence}} & \multicolumn{5}{c}{\textbf{Metrics}} \\
    \cmidrule(lr){1-2} \cmidrule(lr){3-7}
    \textbf{Encoder} & \textbf{Decoder} & \textbf{MoF} & \textbf{Edit} & \multicolumn{3}{c}{\textbf{F1@\{10, 25, 50\}}} \\
    \midrule
    \checkmark & \ding{55} & 35.9 & 37.2 & 34.0 & 27.8 & 16.3 \\
    \ding{55} & \checkmark & 28.1 & 33.9 & 29.8 & 23.9 & 13.7 \\
    \checkmark & \checkmark & \textbf{37.4} & \textbf{39.4} & \textbf{34.7} & \textbf{28.4} & \textbf{16.4}  \\
    \bottomrule
  \end{tabular}}
  \caption{Impact of independent joint embedding in the encoder or decoder on the LARa dataset.\label{tab:asymmetric}}
\end{table}

\subsection{Impact of Initialization}\label{sec:ablation-init}
We initialize the codebook randomly. In Table~\ref{tab:init}, we compare it to the initialization using time series \(k\)-means~\cite{JMLR:v21:20-091}. The results show the model’s robustness to different initialization strategies.

\begin{table}
  \centering
  \setlength{\tabcolsep}{4pt} 
  \begin{tabular}{@{}llllll@{}}
    \toprule
    \textbf{Initialization} &
    \textbf{MoF} &
    \textbf{Edit} &
    \multicolumn{3}{c}{\textbf{F1@\{10, 25, 50\}}} \\
    \midrule
    \textbf{Random} & \textbf{37.4} & 39.4 & 34.7 & \textbf{28.4} & \textbf{16.4}\\

    K-Means & 37.3 & \textbf{40.2} & \textbf{35.2} & 28.2 &  16.0\\

    \bottomrule
  \end{tabular}
  \caption{Impact of initialization on the LARa dataset.\label{tab:init}}
\end{table}

\subsection{Impact of Autoencoder}\label{sec:ablation-autoencoder}
To evaluate the effect of different autoencoder architectures on \textit{\model}, we maintain consistent settings while changing the autoencoder architectures. Specifically, we compare the joint-based disentangled Multi-stage Temporal Convolutional Network (MS-TCN)~\cite{farha2019ms} autoencoder with a Spatial-Temporal Graph Convolutional Network (ST-GCN)~\cite{yan2018spatial} autoencoder. This comparison allows us to assess how each autoencoder architecture influences the performance of our model. Table~\ref{tab:autoencoder} demonstrates that the MS-TCN autoencoder consistently achieves better performance compared to ST-GCN.

\begin{table}
  \centering
  \begin{tabular}{@{\hspace{1em}}l@{\hspace{0.5em}}c@{\hspace{0.5em}}c@{\hspace{0.5em}}ccc@{\hspace{0.5em}}}
    \toprule
    \textbf{Autoencoder} &
    \textbf{MoF} &
    \textbf{Edit}&
    \multicolumn{3}{c}{\textbf{F1@\{10, 25, 50\}}}\\
    \midrule
    ST-GCN & 34.2 & 38.5 & \multicolumn{1}{c}{32.9} & \multicolumn{1}{c}{25.7} & \multicolumn{1}{c}{14.4}\\
    \textbf{MS-TCN} & 
    {\bf 37.4} & 
    \textbf{39.4} &
    \multicolumn{1}{c}{\textbf{34.7}}&
    \multicolumn{1}{c}{\textbf{28.4}}&
    \multicolumn{1}{c}{\textbf{16.4}}\\

    \bottomrule
  \end{tabular}
  \caption{Evaluation of different autoencoders on the LARa dataset.\label{tab:autoencoder}}
\end{table}

\subsection{Impact of Input Skeleton Representation}
In this ablation, we investigated the impact of different input skeleton representations. Specifically, we analyzed how LARa dataset’s features, 3D position coordinates and orientation angles, contribute to the overall performance of \textit{\model}. We conducted experiments where the model was trained with (i) only position coordinates, (ii) only orientation angles, and (iii) a combination of both. The results in Table~\ref{tab:input-features} demonstrate that using both representations yielded the best performance, with position coordinates alone being the second most effective. This indicates that while both position and orientation contribute to the action segmentation, position information plays a more significant role.

\begin{table}
  \resizebox{\columnwidth}{!}{
  \centering
  \setlength{\tabcolsep}{4pt} 
  \begin{tabular}{@{}cclllll@{}}
    \toprule
    \textbf{Position (mm)} &
    \textbf{Orientation (deg)} &
    \textbf{MoF} &
    \textbf{Edit}&
    \multicolumn{3}{c}{\textbf{F1@\{10, 25, 50\}}}\\
    \midrule
    \checkmark & \ding{55} & 33.9 & 39.2 & \textbf{35.3} & 28.1 & 16.3 \\
    \ding{55} & \checkmark & 23.7 & 23.9 & 17.3 & 11.7 & 5.2 \\
    \checkmark & \checkmark & {\bf 37.4} & 
    \textbf{39.4} &
    \multicolumn{1}{c}{34.7}&
    \multicolumn{1}{c}{\textbf{28.4}}&
    \multicolumn{1}{c}{\textbf{16.4}}\\
    \bottomrule
  \end{tabular}}
  \caption{Impact of input skeleton representation on the LARa dataset.\label{tab:input-features}}
\end{table}

\subsection{Impact of Patching}\label{sec:vq_vs_patchvq}
\begin{figure*}[tb]
    \centering
    \begin{subfigure}[b]{0.32\textwidth}
        \centering
        \includegraphics[width=\textwidth,trim={150 0 175 0},clip]{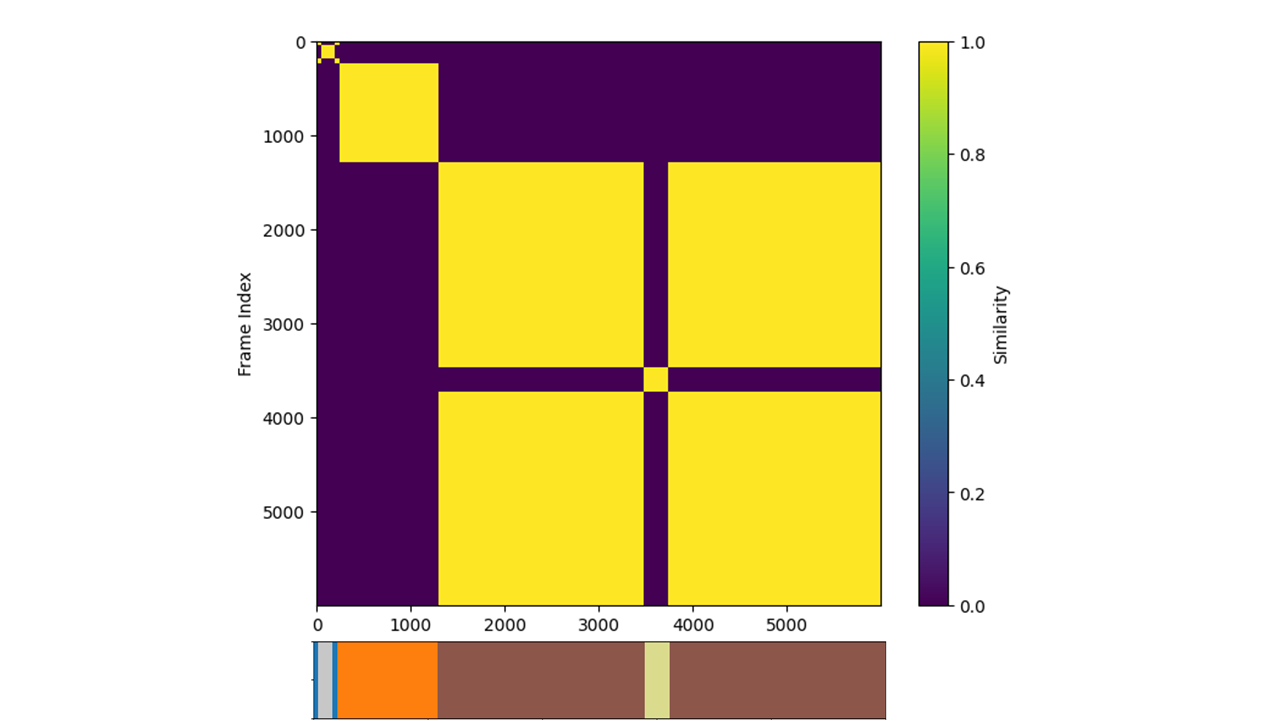}
        \caption{Ground Truth}
        \label{fig:input_features}
    \end{subfigure}
    \begin{subfigure}[b]{0.32\textwidth}
        \centering
        \includegraphics[width=\textwidth,trim={150 0 175 0},clip]{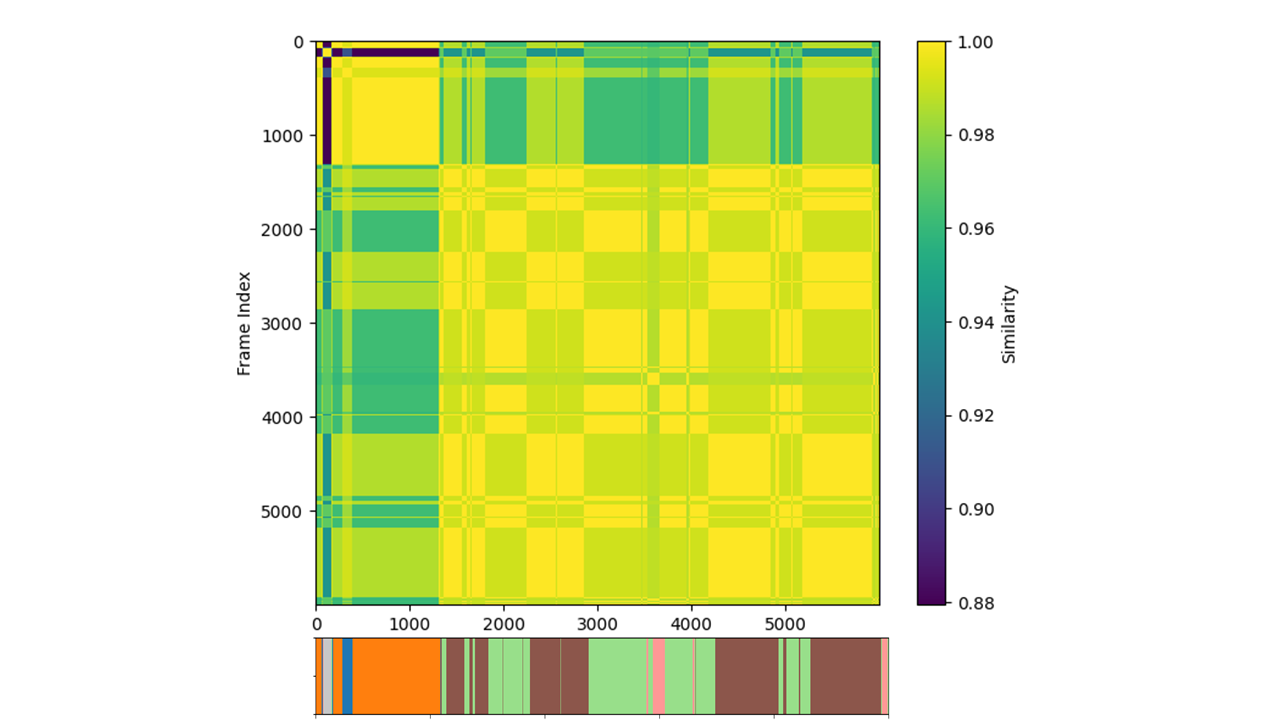}
        \caption{Patch Size 1}
        \label{fig:vq_latent}
    \end{subfigure}
    \begin{subfigure}[b]{0.32\textwidth}
        \centering
        \includegraphics[width=\textwidth,trim={150 0 175 0},clip]{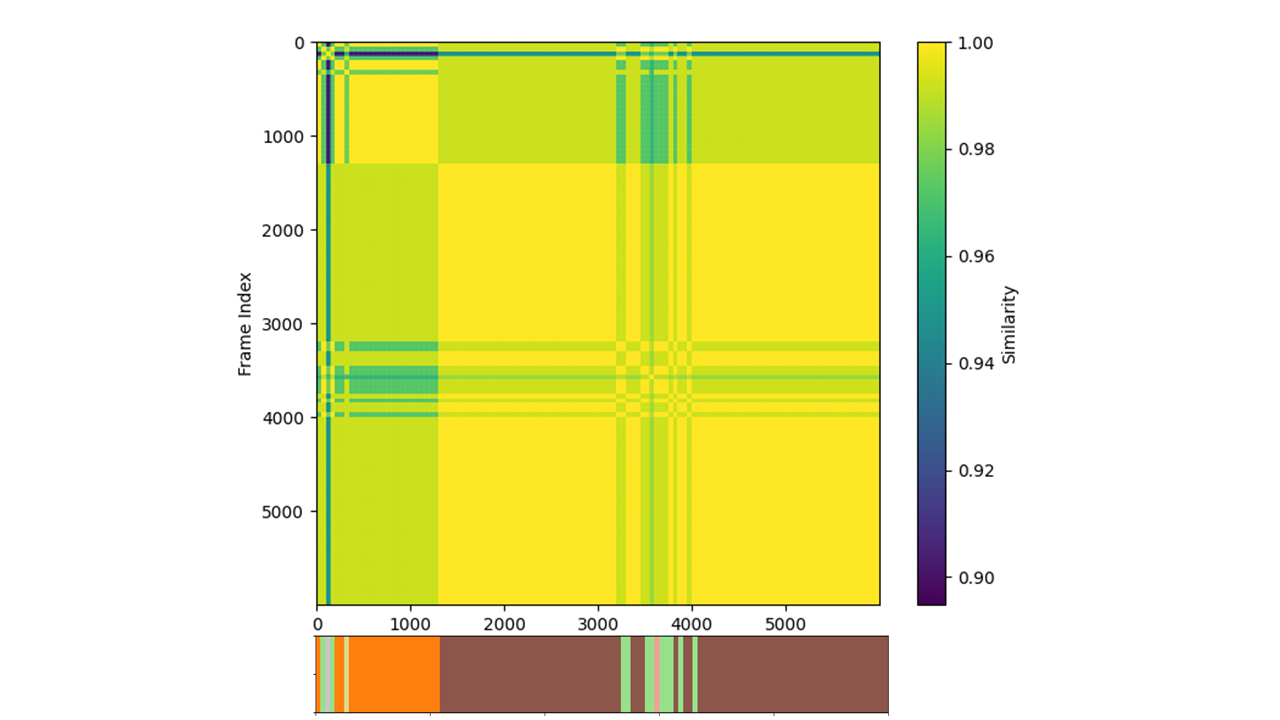}
        \caption{Patch Size 50}
        \label{fig:patchvq_latent}
    \end{subfigure}
    \caption{Comparison of self-similarity matrices for ground truth labels (a), patch size 1 (b) and 50 (c).
    }
    \label{fig:vq_vs_patchvq}
\end{figure*}

We provide a qualitative comparison between the latent representations of \textit{\model} for two different patch sizes in Figure~\ref{fig:vq_vs_patchvq}. We plot the self-similarity matrices of a skeleton sequence from the LARa dataset based on the learned representation for patch size 1 and 50. For patch size 1, the learned representation is quite noisy.

We also evaluate the effectiveness of patch-based processing for other unsupervised temporal action segmentation approaches. To achieve this, we partition the input skeleton features into non-overlapping patches following the latent patching mechanism introduced in our framework. Subsequently, these patches are concatenated into a single elongated vector, which is then fed to the unsupervised action segmentation methods. Note that \textit{\model} performs patching in the latent space, which is not possible with the other methods. Table~\ref{tab:patch_tas_comparison_lara} shows that patch-based processing improves segmental metrics for CTE and TOT. This is likely because predictions of the actions are in patch-level, which have coarser granularity, preventing over-segmentation. However, there is no notable increase in MoF. In contrast, ASOT, which already tends to predict longer segments, shows a decrease in performance under patch-based processing. Even if patch-based processing is added to CTE, TOT, and ASOT, \textit{\model} outperforms them.   

\begin{table}[t]
    \centering
    \resizebox{\columnwidth}{!}{
    \begin{tabular}{@{}l|lcclll@{}}
        \hline
         & \multicolumn{5}{c}{\textbf{LARa}} \\
        \cline{2-6}
        \textbf{Method} & \textbf{MoF} & \textbf{Edit} & \multicolumn{3}{c}{\textbf{F1@\{10, 25, 50\}}} \\
        \hline
         Patch + CTE \cite{kukleva2019unsupervised} & 25.8 & 29.3 & 21.5 & 16.2 & 8.4 \\
         Patch + TOT \cite{kumar2022unsupervised} & 19.3 & 28.1 & 21.4 & 14.4 & 6.4 \\
         Patch + ASOT \cite{xu2024temporally} & 21.1 & 21.5 & 19.4 & 12.6 & 5.2 \\
        \hline
        \textbf{\model\ (ours)} & \textbf{37.4} & \textbf{39.4} & \textbf{34.7} & \textbf{28.4} & \textbf{16.4} \\
        \hline
    \end{tabular}}
    \caption{Evaluation of unsupervised temporal action segmentation methods with patched skeleton input on the LARa dataset.}
    \label{tab:patch_tas_comparison_lara}
\end{table}

\subsection{Runtime}\label{sec:runtime}
Table~\ref{tab:runtime} compares the runtime of various unsupervised temporal action segmentation methods when processing the entire datasets on a computer with a single NVIDIA RTX 4090 GPU.
While the runtime of \textit{\model} is higher compared to CTE~\cite{kukleva2019unsupervised} and ASOT~\cite{xu2024temporally}, all methods are very fast and process entire datasets in less than 80 minutes.

\begin{table}
\centering
\begin{tabular}{cccccc}
\toprule
 & CTE & TOT & ASOT & SMQ \\
\midrule
HuGaDB & 7.8 & 32.5 & 2.2 & 9.8 \\
LARa & 40.2 & 79.8 & 4.3 & 44.0 \\
\bottomrule
\end{tabular}
\caption{Runtime (mins) for the HuGaDB and LARa datasets.\label{tab:runtime}}
\end{table}

\subsection{Visualization of Embeddings}
\begin{figure*}[t]
    \centering
    \begin{subfigure}[b]{0.32\textwidth}
        \centering
        \includegraphics[width=\textwidth]{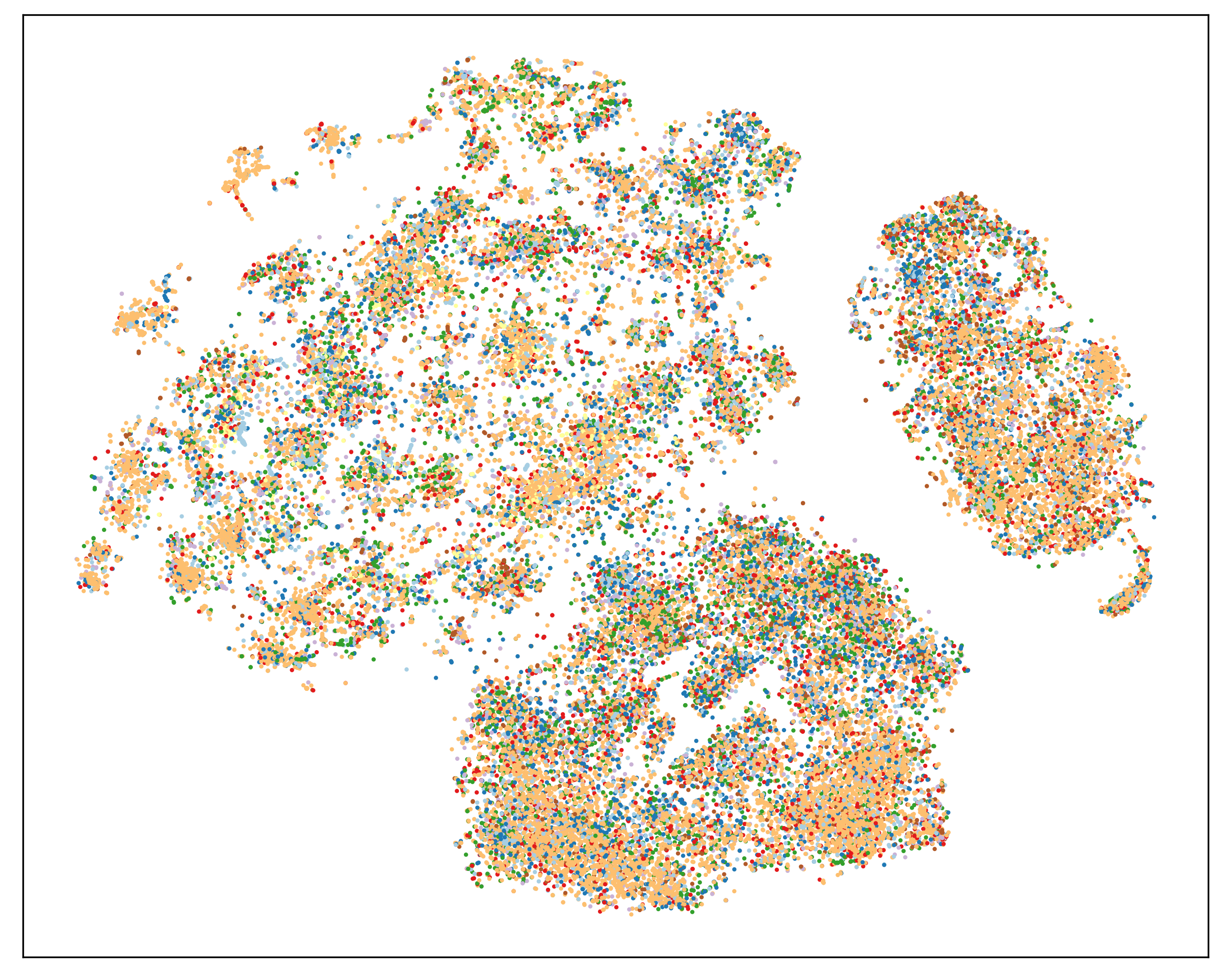}
        \caption{Input Features}
        \label{fig:input_features_tsne}
    \end{subfigure}
    \begin{subfigure}[b]{0.32\textwidth}
        \centering
        \includegraphics[width=\textwidth]{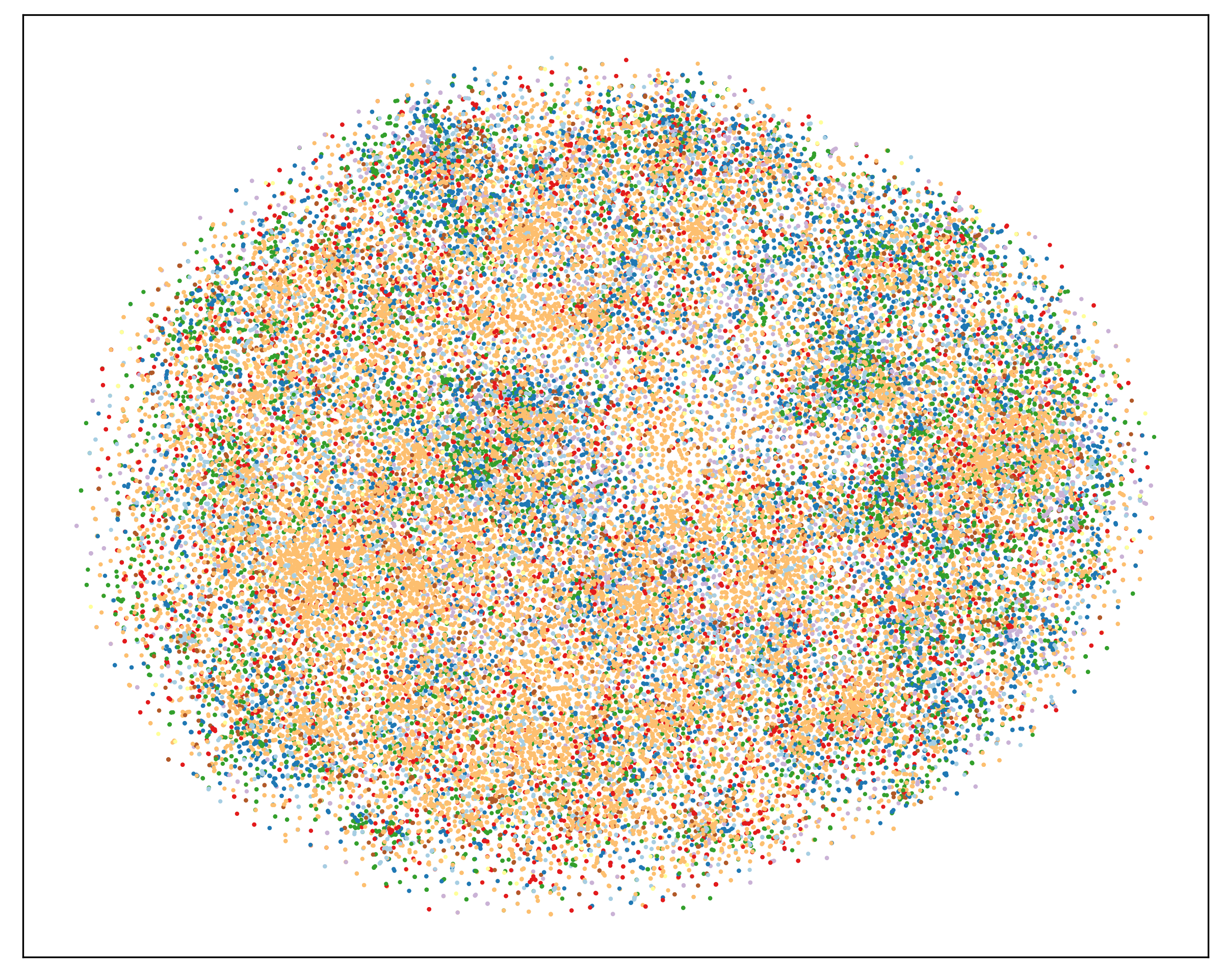}
        \caption{Predict\&Cluster}
        \label{fig:predict_cluster_tsne}
    \end{subfigure}
    \begin{subfigure}[b]{0.32\textwidth}
        \centering
        \includegraphics[width=\textwidth]{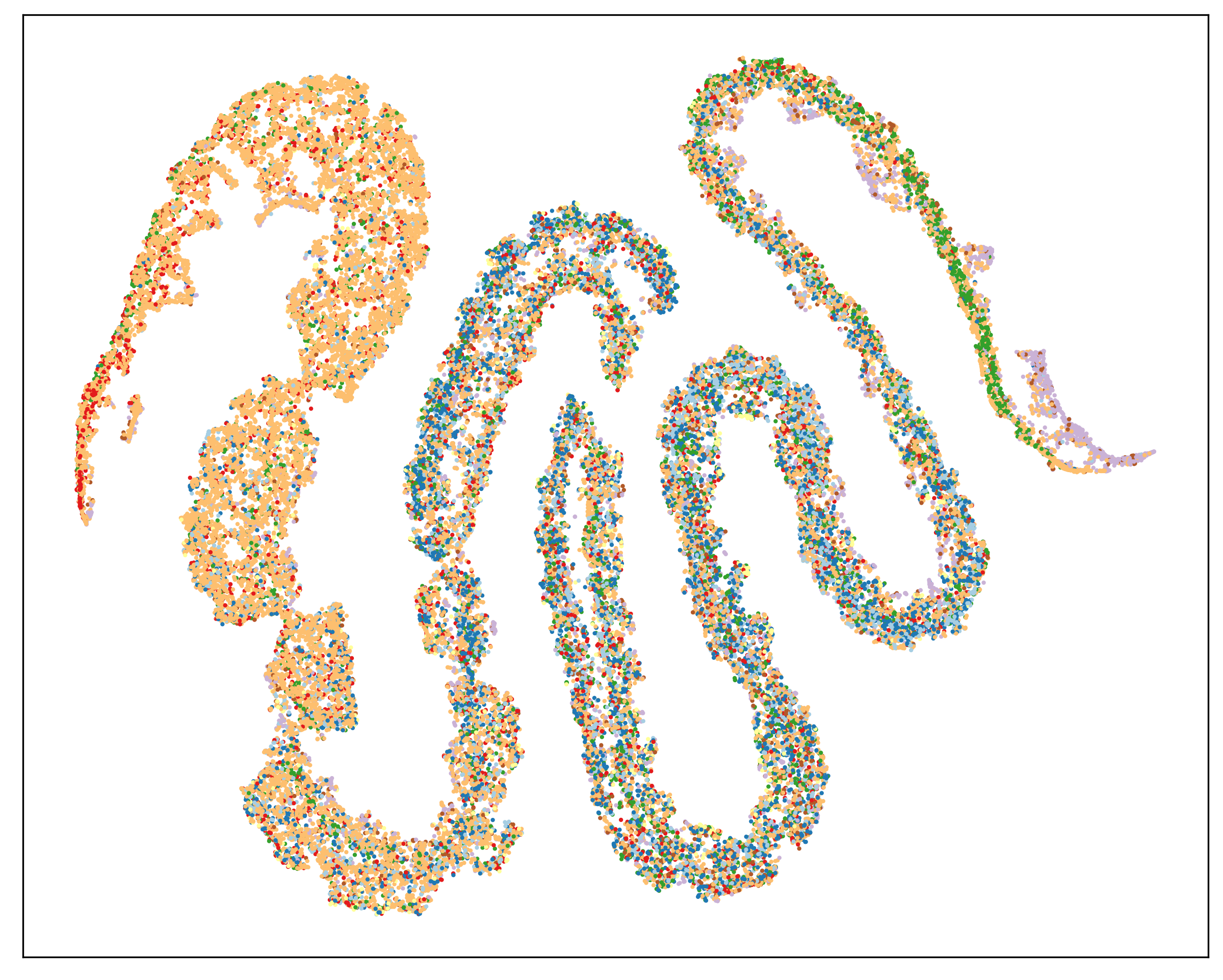}
        \caption{AimCLR}
        \label{fig:aimclr_tsne}
    \end{subfigure}
    \begin{subfigure}[b]{0.32\textwidth}
        \centering
        \includegraphics[width=\textwidth]{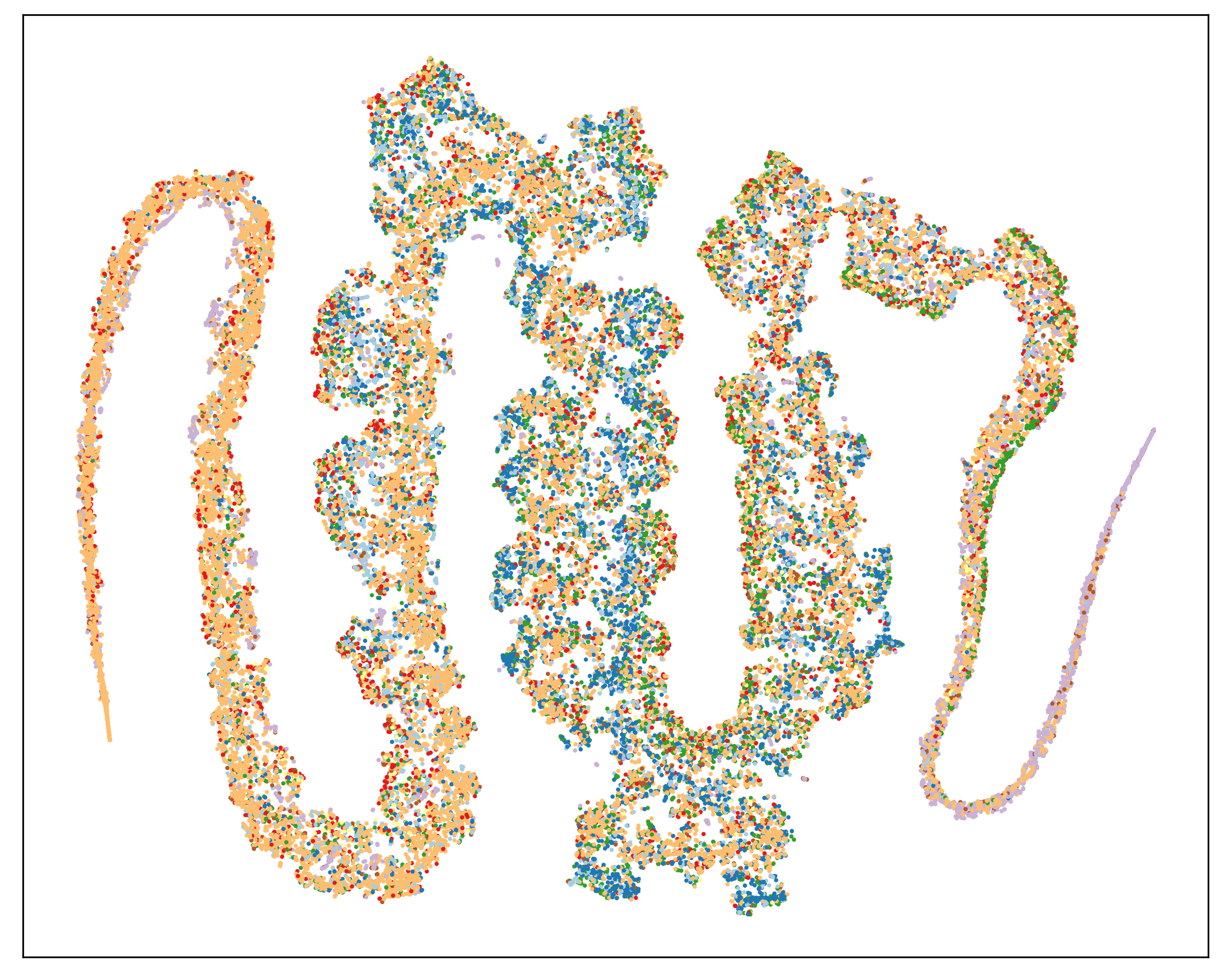}
        \caption{ActCLR}
        \label{fig:actclr_tsne}
    \end{subfigure}
    \begin{subfigure}[b]{0.32\textwidth}
        \centering
        \includegraphics[width=\textwidth]{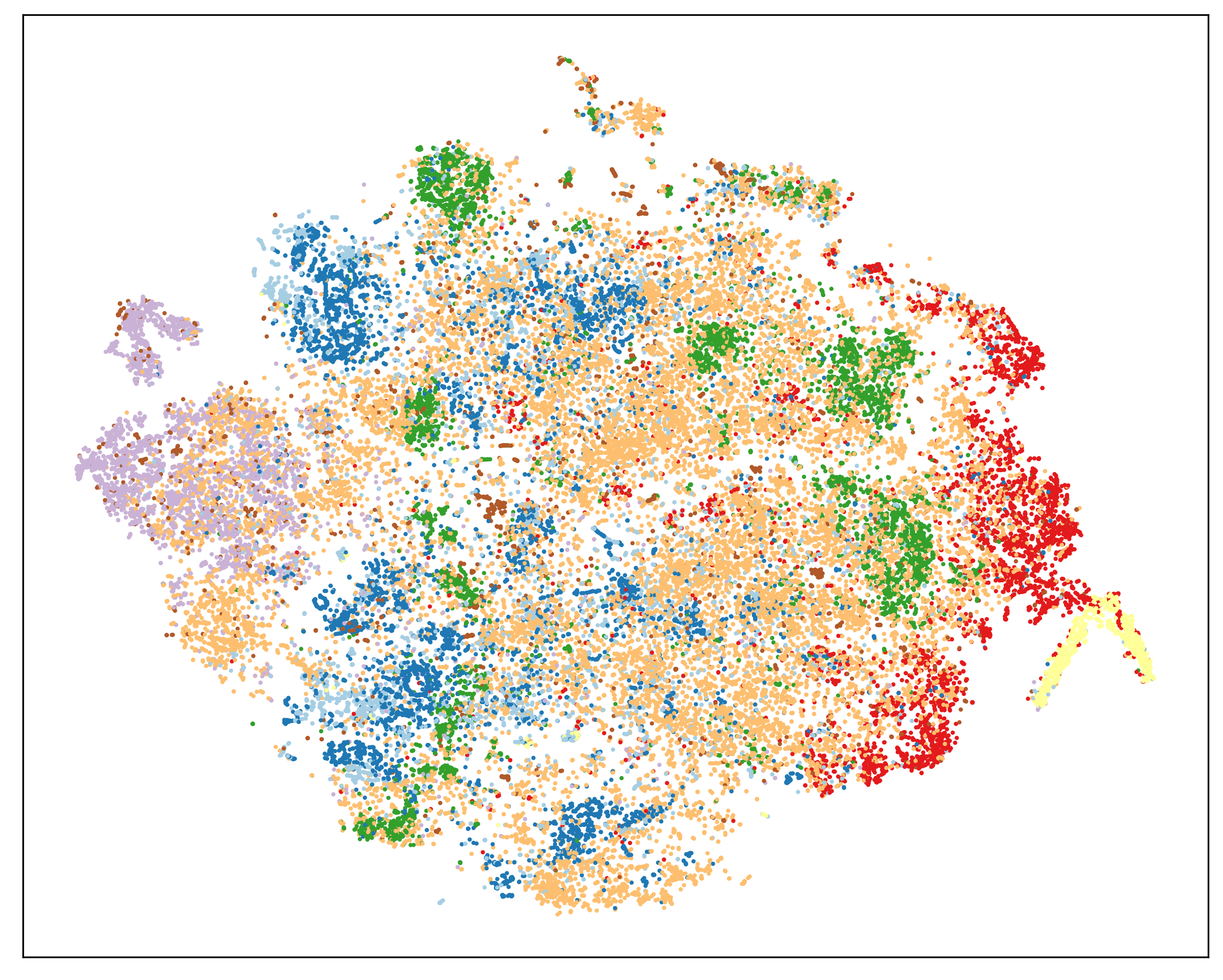}
        \caption{\model\ (Ours)}
        \label{fig:patchvq_tsne}
    \end{subfigure}
    \caption{The t-SNE visualizations of the input skeleton features and latent embeddings generated by various methods on the LARa dataset. Each point corresponds to a patch, and the colors represent the ground truth action labels.}
    \label{fig:tsne_patchvq}
\end{figure*}

Furthermore, we visualize latent embeddings using t-SNE in Figure~\ref{fig:tsne_patchvq}. We color each point based on ground-truth labels, and each point represents a latent patch. The plots reveal that \textit{\model} produces a more distinctive action representation compared to other methods.

\subsection{Impact of Number of Actions (K)}\label{sec:ablation-num_k}
\begin{figure}[t]
    \centering
    \begin{subfigure}{0.5\textwidth}
        \centering
        \includegraphics[width=\textwidth]{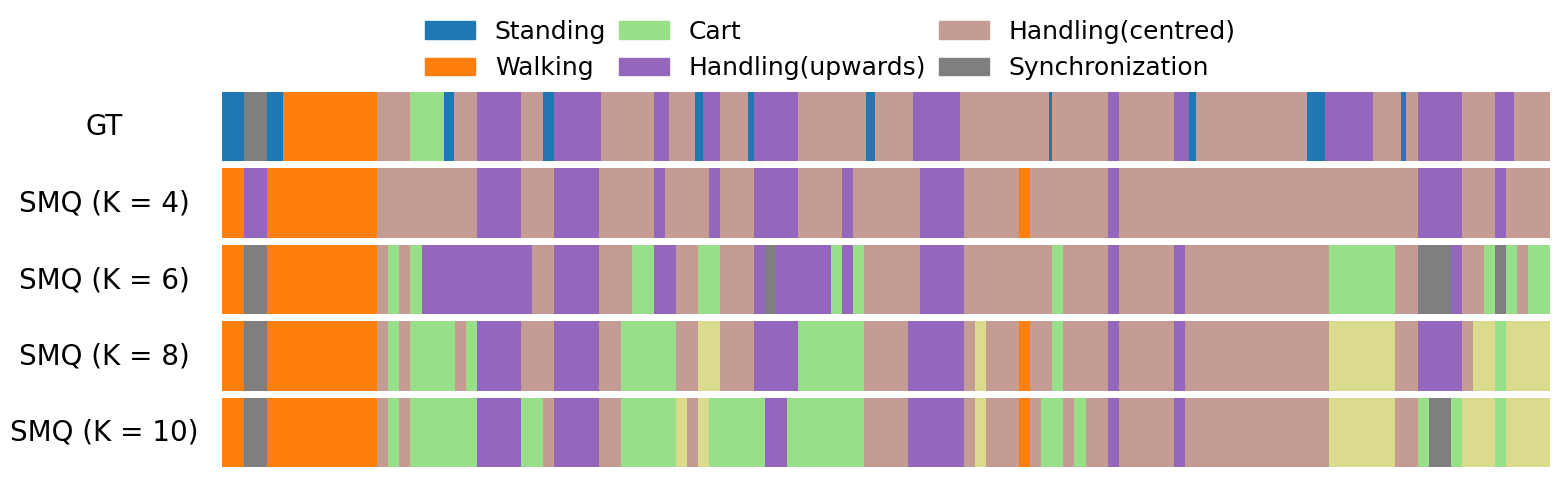}
        \label{fig:subfigure1-k}
    \end{subfigure}
    \hfill
    \begin{subfigure}{0.5\textwidth}
        \centering
        \includegraphics[width=\textwidth]{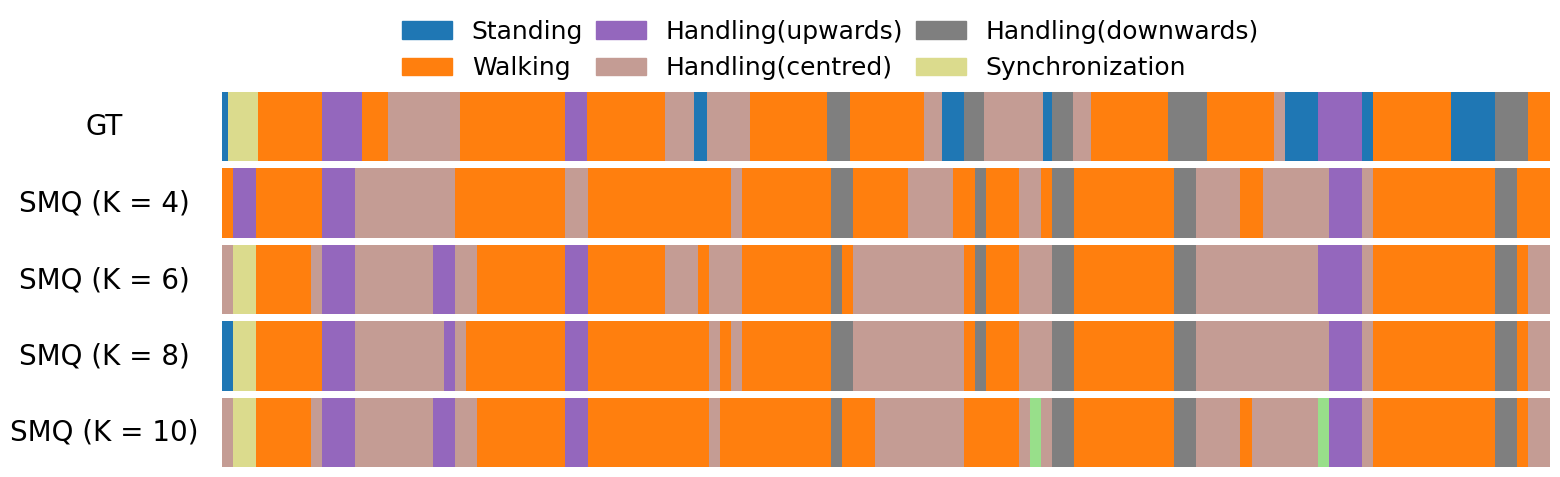}
        \label{fig:subfigure2-k}
    \end{subfigure}
    \caption{Qualitative results for \model\ with different values of $K$ for the LARa dataset.}
    \label{fig:ablation-k}
\end{figure}

Providing the ground truth number of action classes (K) is standard in the evaluation of unsupervised video-based temporal action segmentation methods \cite{kukleva2019unsupervised, kumar2022unsupervised, xu2024temporally}. Accordingly, we used this procedure in our experiments.
To assess our model's robustness to variations in K, we conducted an ablation study by systematically changing the provided number of action classes on the LARa dataset as shown in Table~\ref{tab:varying_k}. Our results demonstrate that the model consistently maintains strong performance despite changes in K, highlighting its robustness. Figure \ref{fig:ablation-k} shows some qualitative results for different values of K. Note that a smaller value of K discovers less actions.

\begin{table}[t]
  \centering
  \setlength{\tabcolsep}{4pt} 
  \begin{tabular}{@{}llllll@{}}
    \toprule
    \textbf{Num of actions (K)} &
    \textbf{MoF} &
    \textbf{Edit} &
    \multicolumn{3}{c}{\textbf{F1@\{10, 25, 50\}}} \\
    \midrule
    3 & 41.8 & 40.9 & 41.7 & 33.9 & 20.0 \\
    4 & 44.0 & 42.2 & 41.2 & 35.3 & 21.7 \\
    5 & 41.9 & 39.0 & 37.6 & 31.2 & 18.0 \\
    6 & 38.2 & 39.4 & 36.7 & 29.8 & 17.3 \\
    7 & 37.4 & 41.2 & 35.8 & 29.1 & 16.8 \\
    \hline
    8 (GT) & 37.4 & 39.4 & 34.7 & 28.4 & 16.4 \\
    \hline
    9 & 32.7 & 39.4 & 34.7 & 27.9 & 16.3 \\
    10 & 34.0 & 36.8 & 33.5 & 27.0 & 15.5 \\
    \bottomrule
  \end{tabular}
  \caption{Effect of varying K for the LARa dataset.\label{tab:varying_k}}  
\end{table}

\begin{figure}[ht]
  \centering
  \includegraphics[width=\columnwidth]{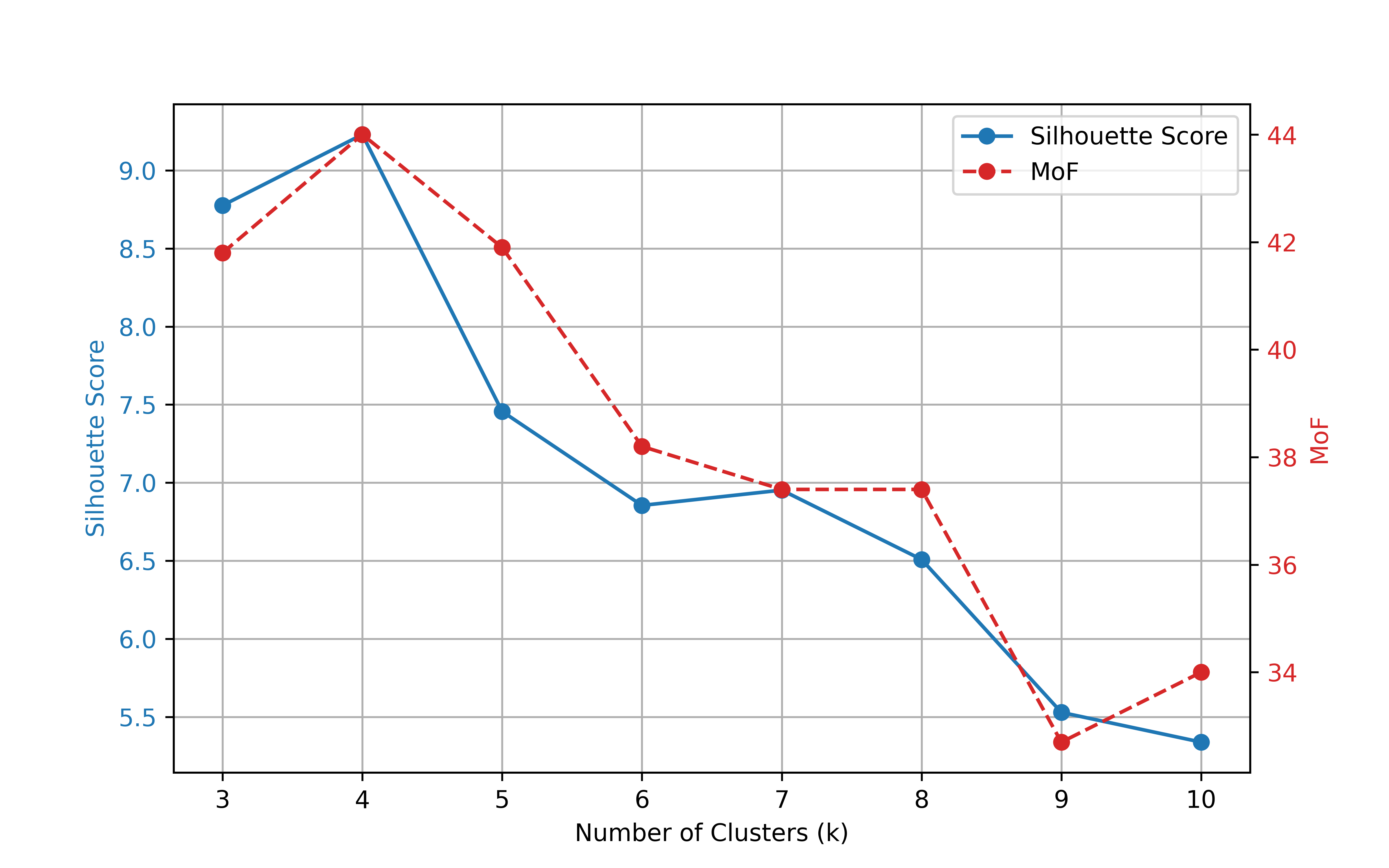}
  \caption{Silhouette score and MoF for varying $K$ on the LARa dataset.}
  \label{fig:Silhouette}
\end{figure}

In practice, the value of K can be determined by the silhouette score, which quantifies both cohesion within clusters and separation between them. The silhouette score is calculated based on the latent patch embeddings and their corresponding cluster assignments. Figure~\ref{fig:Silhouette} shows the patch-based silhouette score~\cite{JMLR:v21:20-091} and MoF for different K values. Since the silhouette score is highly correlated with MoF, it can be used to determine K.

\subsection{Sequence-level Temporal Action Segmentation}
While our approach is designed to discover and segment actions across entire datasets, it can also be evaluated on individual sequences. To ensure a fair comparison, we apply local Hungarian matching on each skeleton sequence separately, determining the best ground-truth-to-cluster-label mapping per sequence. Sequence-level temporal action segmentation methods~\cite{sarfraz2019efficient, sarfraz2021tw-finch, bueno2023tsa} require the number of actions per sequence, so the average number of unique actions per sequence is calculated over the entire dataset and provided. The results are reported in Table~\ref{tab:tas_comparison-local-hungarian}. \textit{\model} outperforms sequence-level segmentation approaches, even though these techniques benefit from local Hungarian matching based on the average number of actions per sequence. In contrast, our approach only requires the total number of actions $K$ in the dataset, not per sequence.

\begin{table*}[tb]
    \centering
    
    \begin{tabular}{@{}l|lllll|lllll@{}}
        \hline
         & \multicolumn{5}{|c|}{\textbf{HuGaDB}} & \multicolumn{5}{c}{\textbf{LARa}} \\
        \cline{2-11}
        \textbf{Method} & \textbf{MoF} & \textbf{Edit} & \multicolumn{3}{c|}{\textbf{F1@\{10, 25, 50\}}} & \textbf{MoF} & \textbf{Edit} & \multicolumn{3}{c}{\textbf{F1@\{10, 25, 50\}}} \\
        \hline
        FINCH \cite{sarfraz2019efficient} & 54.2 & 19.5 & 5.9 & 3.6 & 3.1 & 43.7 & 38.0 & 36.2 & 29.1 & 17.6 \\
        TW-FINCH \cite{sarfraz2021tw-finch} & 57.5 & 39.8 & 45.8 & 36.4 & 25.8 & 37.5 & 17.8 & 22.7 & 12.2 & 4.0 \\
        TSA (K-means) \cite{bueno2023tsa} & 58.0 & 22.8 & 10.6 & 7.3 & 5.2 & 36.6 & 18.5 & 22.1 & 12.3 & 4.8 \\
        TSA (FINCH) \cite{bueno2023tsa} & \textbf{61.6} & 37.3 & 17.1 & 13.8 & 10.2 & 39.8 & 16.3 & 22.0 & 12.7 & 5.2 \\
        TSA (TW-FINCH) \cite{bueno2023tsa} & 54.1 & 38.5 & 37.4 & 27.4 & 17.6 & 36.4 & 14.1 & 22.6 & 11.8 & 4.2 \\
        \hline
        \textbf{\model\ (ours)} & 55.3 & \textbf{44.5} & \textbf{52.8} & \textbf{44.6} & \textbf{31.7} & 
        \textbf{51.7} & \textbf{45.1} & \textbf{45.3} & \textbf{38.8} & \textbf{23.6} \\
        \hline
    \end{tabular}
    \caption{Human motion segmentation (single sequence) on the HuGaDB and LARa datasets.}
    \label{tab:tas_comparison-local-hungarian}
\end{table*}


\subsection{Robustness to Missing Joints}
To assess robustness to missing joints, we performed an ablation study on the LARa dataset by randomly dropping 25\% (5 joints) and 50\% (11 joints) of the 22 full-body joints using three different random seeds. We trained and evaluated the model under each setting and averaged the results. Additionally, we evaluated a structured setting where only the wrist and hand joints from both arms were removed. As shown in Table~\ref{tab:ablation-missing-joint}, randomly dropping joints did not lead to a substantial performance drop, demonstrating SMQ's robustness to incomplete or degraded skeleton data. However, removing the wrist and hand joints resulted in a more noticeable decline, suggesting that these joints carry semantically important cues for distinguishing actions.

\begin{table}[tb]
  \centering
  \resizebox{\columnwidth}{!}{
  \setlength{\tabcolsep}{4pt}
  \begin{tabular}{@{}llllll@{}}
    \toprule
    \textbf{Method} &
    \textbf{MoF} &
    \textbf{Edit} &
    \multicolumn{3}{c}{\textbf{F1@\{10, 25, 50\}}} \\
    \midrule
    25\% joints missing (17 joints) & 36.0 & 39.2 & \textbf{34.9} & 28.1 & 16.1 \\
    50\% joints missing (11 joints) & 34.8 & 38.7 & 33.9 & 27.2 & 15.5 \\
    Hand and wrist missing (18 joints) & 33.1 & 37.0 & 34.3 & 26.3 & 14.9 \\
    All joints (22 joints) & \textbf{37.4} & \textbf{39.4} & 34.7 & \textbf{28.4} & \textbf{16.4} \\
    \bottomrule
  \end{tabular}
  }
  \caption{Investigation of SMQ's robustness to missing joints.\label{tab:ablation-missing-joint}}
\end{table}

\end{document}